\documentclass[journal]{IEEEtai}

\usepackage[colorlinks,urlcolor=blue,linkcolor=blue,citecolor=blue]{hyperref}

\usepackage{color,array}

\usepackage{graphicx}

\usepackage{algorithm}
\usepackage{algpseudocode}
\usepackage{amsmath,amssymb}
\usepackage{booktabs}
\usepackage{listings}
\usepackage{cite}
\usepackage{url}
\usepackage{comment}
\usepackage{subcaption}
\usepackage{siunitx}
\usepackage{multirow}
\usepackage{threeparttable}
\usepackage{balance}
\usepackage{pdfpages} 

\DeclareSIUnit{\tok}{tok}
\DeclareSIUnit{\flop}{FLOP}

\begin{document}

\title{Profiling Lightweight Large Language Models}

\author{
Tomohiro~Harada,~\IEEEmembership{Member,~IEEE,}
Enrique~Alba,~\IEEEmembership{Member,~IEEE,}
and~Gabriel~Luque,~\IEEEmembership{Member,~IEEE}
\thanks{T. Harada is with the Graduate School of Science and Engineering, Saitama University, Saitama, Japan e-mail: tharada@mail.saitama-u.ac.jp.}
\thanks{E. Alba and G. Luque are with ITIS, University of Malaga, 29071 Malaga, Spain e-mail: (ealbat@uma.es, gluque@uma.es).}
\thanks{Code and data is available at \url{https://github.com/tharada0126/llm_ptme.git}.}
\thanks{Preprint.}}

\markboth{Preprint}%
{T. Harada \MakeLowercase{\textit{et al.}}: Profiling Lightweight Large Language Models}

\maketitle

\begin{abstract}
Lightweight large language models (LLMs) are increasingly being deployed locally on personal computers and are expected to play a growing role in resource-constrained edge and mobile environments. In such settings, energy consumption, execution time, and memory usage directly affect practical usability, yet existing evaluations of LLM efficiency largely rely on proxy descriptors such as parameter count or FLOPs, often decoupled from task precision. This paper introduces a PTME-based experimental framework for the precision-aware profiling of lightweight LLM inference, jointly measuring Precision, execution Time, peak Memory usage, and Energy consumption through direct hardware-level measurements. The methodology is applied to a representative set of lightweight LLMs executed locally under edge-class resource envelopes on a controlled desktop platform, using benchmarks spanning code generation, mathematical reasoning, and multi-task understanding. We find that static proxy descriptors approximate inference cost well but fail to predict precision. Tightening the resource envelope increases cost without affecting precision, amplifying execution time more strongly than energy and penalizing larger models the most. 
Moreover, no single model dominates across all PTME dimensions, and a Pareto analysis reveals non-dominated configurations that would be hidden by accuracy-only or efficiency-only assessments, providing practical guidance for selecting models under different resource envelopes. These results show that selecting lightweight LLMs by size, FLOPs, latency, or accuracy alone can select the wrong deployment candidate; PTME profiling exposes configurations that preserve useful accuracy at lower physical cost. 

\end{abstract}

\begin{IEEEImpStatement}
Large language models are increasingly run on personal and edge devices, where energy, time, and memory are scarce and directly limit what is practical. Today, these models are often chosen using indirect indicators such as parameter count, which this work shows can be misleading: such indicators track running costs well but say nothing about whether a model is actually accurate. By measuring accuracy against real energy, time, and memory usage on constrained hardware, this work provides developers with a practical, evidence-based way to select models that are both capable and efficient, rather than merely small. This matters economically by avoiding wasted compute and over-provisioned hardware, and environmentally by reducing the energy and carbon costs of everyday AI use as inference moves closer to billions of end-user devices.
\end{IEEEImpStatement}

\section{Introduction}
\label{sec:introduction}
Large language models (LLMs) have become a foundational component of modern artificial intelligence (AI), enabling high-quality natural language understanding, reasoning, and code generation across a broad spectrum of applications. As LLMs transition from experimental systems to everyday computational tools, their inference workload increasingly dominates the operational cost of AI deployments. This trend raises significant concerns regarding energy consumption, environmental impact, and sustainability, particularly as inference requests scale in volume and move closer to end users~\cite{strubell2019energy, schwartz2020green}. While large cloud infrastructures can partially amortize energy costs, such assumptions no longer hold as LLM inference increasingly shifts toward personal computers and resource-constrained edge and mobile environments.

This shift toward local deployment is driven by latency requirements, privacy constraints, offline availability, and cost considerations~\cite{Abadade2023TinyML,xu2024ondevice,KIBRIYA2024109698}. Users increasingly expect interactive LLM-based functionality on laptops and desktops without relying on constant connectivity to remote servers. At the same time, emerging use cases on light and embedded devices, such as smartphones, single-board computers (e.g., Raspberry Pi), and microcontroller-adjacent platforms, further tighten the feasible execution envelope. In these environments, energy availability, thermal envelopes, CPU parallelism, and memory capacity fundamentally constrain feasible model sizes and inference strategies~\cite{shi2016edge, satyanarayanan2017edge}. Consequently, energy-efficient inference is not merely an optimization objective but a prerequisite for practical deployment.

Recent work has explored techniques to reduce the inference cost of LLMs, including quantization, pruning, and lightweight model architectures~\cite{dettmers2022llm, frantar2023gptq,NEURIPS2023_44956951,Zheng2025Review}. These approaches have demonstrated substantial reductions in memory usage and latency and, in some cases, lower energy consumption. However, model efficiency is still often assessed using static proxy descriptors such as parameter count, theoretical FLOPs, or loaded model size, or using isolated runtime measurements such as latency alone. Such descriptors are useful for preliminary screening, but they cannot fully characterize deployment behavior on real hardware, where memory access patterns, cache utilization, backend implementations, numerical precision, and resource constraints interact~\cite{horowitz2014energy}. More importantly, static proxies do not directly indicate whether a model is accurate enough for the target task. As a result, models that appear efficient according to proxy metrics may still be unsuitable when precision, time, memory, and energy are considered jointly.

Moreover, energy efficiency is often evaluated independently of task correctness. In practice, reducing energy consumption at the expense of unacceptable accuracy renders an LLM deployment ineffective, particularly in interactive or autonomous settings. Despite this, the literature lacks a unified evaluation framework that jointly considers \emph{accuracy}, \emph{execution time}, \emph{memory footprint}, and \emph{energy consumption} under realistic inference conditions on personal computers and edge-class resource envelopes. Existing studies typically focus on one or two dimensions in isolation, making it difficult to reason about trade-offs or determine whether a model is genuinely efficient for a target deployment scenario.

In previous works, the \emph{PTME} framework, which integrates Precision (Accuracy), Time, Memory, and Energy, has been introduced in related contexts such as energy-aware metaheuristics~\cite{alba2026energyawaremetaheuristics}, machine-learning surrogate modeling~\cite{HARADA2026108482}, and software engineering~\cite{alba2026software}, where solution quality is evaluated together with physical resource consumption. However, its use as a deployment-oriented evaluation framework for lightweight LLM inference remains underexplored. This gap is important because LLM deployment decisions require not only estimating physical costs but also determining whether those costs are justified by task precision in the target environment.

The goal of this paper is to demonstrate that PTME is necessary for the meaningful assessment of lightweight LLMs on personal and edge devices. To this end, we introduce the PTME framework into a unified experimental methodology grounded in direct hardware measurements. Our study is guided by the following research questions:

\begin{itemize}
\item \textbf{RQ1:} To what extent do commonly used static proxy descriptors (e.g., parameter count, FLOPs, loaded memory) correlate with the \emph{measured} PTME metrics of LLM inference on resource-constrained hardware?
\item \textbf{RQ2:} How do precision and resource cost trade off across models, and which configurations are non-dominated and most appropriate for selection?
\item \textbf{RQ3:} How does tightening the resource envelope across different deployment scenarios affect the measured PTME metrics, and how does this effect depend on model scale and task?
\end{itemize}

To answer these questions, we conduct a systematic experimental study of multiple lightweight LLMs executed locally on bounded-resource configurations, using direct energy measurements and diverse inference workloads. The contributions of this work are threefold: (i) we formalize PTME metrics that explicitly integrate accuracy with physical resource consumption; (ii) we empirically show that static proxy descriptors provide useful first-order estimates of inference cost but fail to predict precision; and (iii) we provide reproducible experimental protocols and Pareto-based data analysis that reveal how precision, time, memory, and energy interact across models and resource envelopes, thereby translating profiling results into practical model-selection guidance for sustainable AI deployment.

The remainder of this paper is organized as follows. Section~\ref{sec:models} describes the lightweight LLMs evaluated in this study. Section~\ref{sec:dataset} presents the three benchmarks and their accuracy metrics. Section~\ref{sec:ptme_metrics} defines the PTME metrics. Section~\ref{sec:methodology} details the experimental methodology, and Section~\ref{sec:design} describes the experimental design, including the execution environment, resource envelopes, and evaluated models. Section~\ref{sec:results} presents and analyzes the results with respect to the three research questions. Finally, Section~\ref{sec:conclusion} concludes with implications and directions for future work.

\section{LLMs Under Study}
\label{sec:models}

The evaluated models are selected to represent the current generation of lightweight LLMs that can be executed locally under bounded compute and memory resources without distributed inference. Recent work on efficient and small language models highlights the 1B--7B parameter regime as a practical range for single-node execution while retaining strong task performance~\cite{touvron2023llama,jiang2023mistral,zhang2024tinyllama}. This scale is particularly relevant to local and resource-constrained deployment because it covers models small enough to run on personal computers and edge-class resource envelopes, while still spanning a meaningful range of inference cost and capabilities.

The selected models include TinyLlama-1.1B \cite{zhang2024tinyllama}, Qwen2.5-1.5B \cite{qwen2025qwen25technicalreport}, Gemma-2B \cite{gemma2024technical}, Phi-2 (2.7B) \cite{phi2_microsoft}, Phi-3-mini \cite{abdin2024phi3}, and Mistral-7B \cite{jiang2023mistral}. TinyLlama-1.1B provides a compact reproduction of Llama-style architectures at reduced scale. Qwen2.5 and Gemma represent compact instruction-tuned transformer architectures designed for efficient local inference. The Phi family emphasizes strong reasoning capability per parameter at moderate model sizes. Mistral-7B serves as the largest reference model in our study, providing an upper-bound point within the compact-model regime commonly used in local execution environments. All evaluated models use quantized variants to reflect practical local deployment scenarios. 

Together, these models cover multiple model families, parameter scales, and tokenizer implementations while remaining executable within the constrained resource envelopes considered in this study. This diversity is important because inference cost and task precision are not determined solely by parameter count; they may also depend on architectural choices, tokenizer behavior, and backend implementation. To ensure a fair comparison, all models are executed with the same inference backend, and each model is evaluated with its default quantization variant provided by Ollama, without additional post-processing. Thus, the comparisons reflect realistic deployment conditions. Detailed model configurations are summarized in the Supplementary material.

\section{Datasets and Accuracy Metrics}
\label{sec:dataset}

To evaluate lightweight LLMs across diverse workloads and avoid task-specific bias, we select three complementary benchmarks spanning distinct task domains: code generation (engineering), mathematical reasoning, and broad multi-task understanding (general knowledge). The three domains differ in the type of reasoning required, the structure of their outputs, and their prompt and generation lengths, allowing us to examine precision and resource trade-offs under markedly different workloads. At the same time, all three provide objective, deterministic correctness criteria, enabling reproducible measurement of computational and energy behavior, in contrast to open-ended generation tasks whose correctness is hard to quantify. Although the selected benchmarks are not exhaustive, they provide representative workloads for evaluating lightweight LLM inference. More importantly, the proposed PTME evaluation framework is independent of the specific benchmark and can be applied to a wide range of tasks. This selection also aligns with recent empirical studies on efficient LLM evaluation and sustainable inference, which emphasize benchmarking under realistic and diverse workloads to understand computational and energy trade-offs~\cite{henderson2020sustainable,schwartz2020green}.

\subsection{HumanEval}

HumanEval (HE)~\cite{chen2021evaluatinglargelanguagemodels} is a widely adopted benchmark for evaluating functional code generation through unit-test-based verification. Each instance consists of a short natural-language problem specification and a function signature, and correctness is determined by executing the generated function against hidden test cases. Prompt length is relatively controlled, allowing isolation of pure code synthesis behavior without repository context or multi-file reasoning. For this study, we adopt a fixed prompt template that presents the problem description and function signature verbatim as defined in the benchmark, without contextual augmentation or restructuring. Precision $P_{\text{HE}}$ is computed as pass@1 functional correctness, following the standard HumanEval protocol.

\subsection{GSM8K}

GSM8K (GS)~\cite{cobbe2021trainingverifierssolvemath} is a benchmark of grade-school mathematical reasoning problems requiring multi-step symbolic computation. Each instance consists of a natural-language problem and a short numerical answer. Correctness is determined by exact match of the final numerical output. Prompts are constructed directly from the benchmark problem statements without additional contextual information or external knowledge augmentation. GSM8K features moderate input lengths and short outputs, making it suitable for analyzing reasoning-focused inference behavior under constrained resource configurations. Precision $P_{\text{GSM}}$ is defined as exact match on the final numerical answer.

\subsection{MMLU-Pro}

MMLU-Pro (MP)~\cite{Wang2024MMLU-Pro} is a challenging multi-task language understanding benchmark designed to evaluate the reasoning ability of LLMs across 14 broad academic domains. It extends the original MMLU~\cite{hendrycks2021MMLU} by incorporating more reasoning-intensive questions and increasing the number of answer choices from four to ten, thereby reducing the effect of random guessing and improving the discriminative power of the benchmark. Each instance consists of a natural-language question and multiple-choice options, and correctness is determined by whether the model selects the correct answer option. We evaluate MMLU-Pro in a zero-shot setting using the official prompt format without additional demonstrations. Precision $P_{\text{MP}}$ is defined as multiple-choice answer accuracy.

\section{PTME Metrics}
\label{sec:ptme_metrics}

All experiments are evaluated using PTME metrics, which are measured directly during inference execution or derived from benchmark-defined evaluation procedures. Metrics are reported independently and analyzed jointly; no aggregation into a single scalar score is assumed.

Precision ($P$) denotes benchmark correctness and is defined according to the dataset under consideration. For each configuration, precision is computed as
\begin{equation}
P = \frac{1}{N} \sum_{i=1}^{N} s_i,
\end{equation}
where $s_i \in \{0,1\}$ indicates whether the $i$-th instance is solved correctly according to the benchmark criterion (e.g., pass@1 for HumanEval, exact-match for GSM8K and correct option for MMLU-Pro), and $N$ is the number of evaluated instances.

Time ($T$) corresponds to the end-to-end inference time and is measured as wall-clock elapsed time:
\begin{equation}
T = t_{\text{end}} - t_{\text{start}},
\end{equation}
including both prompt processing and autoregressive decoding. 

Memory ($M$) is quantified as the peak resident set size (RSS) during execution:
\begin{equation}
M = \max_{t \in [t_{\text{start}},\, t_{\text{end}}]} \text{RSS}(t),
\end{equation}
where $\text{RSS}(t)$ is the resident set size of the inference backend process and its children together with the profiling process, capturing both static footprint and dynamic allocations.

Energy ($E$) is measured using Intel RAPL CPU-package counters and is computed as the energy consumed by the CPU package during inference:
\begin{equation}
E = E_{\text{pkg}}(t_{\text{end}}) - E_{\text{pkg}}(t_{\text{start}}),
\end{equation}
where $E_{\text{pkg}}(t)$ denotes the cumulative CPU package energy reported by the RAPL interface at time $t$.

For each experimental configuration, PTME metrics are aggregated over the evaluated benchmark instances. These measurements are subsequently used to analyze the correlation between static proxies and the measured PTME metrics (RQ1), the trade-off between precision and resource cost across models under a fixed envelope (RQ2), and the sensitivity of the PTME metrics to the resource envelope (RQ3).

\section{Experimental Methodology}
\label{sec:methodology}

This section describes the experimental methodology used to evaluate lightweight LLMs. We jointly measure accuracy, inference time, peak memory usage, and energy consumption, while enforcing a strict separation between resource measurement and accuracy evaluation. All experiments are executed locally on resource-bounded hardware using a uniform inference backend to ensure comparability across models.

\subsection{Measurement Procedure}
Each experimental configuration is defined by the tuple $(\text{model}, \text{dataset}, \text{resource envelope}, \text{decoding config})$. Decoding configuration consists of the decoding parameters used during inference (temperature, top-p, top-k, max tokens, stop sequences, etc.). For each configuration, we evaluate a fixed subset of benchmark instances, executing inference once per instance, and aggregate the per-instance measurements. To reduce measurement noise caused by operating system scheduling, thermal effects, and background activity, we discard an initial warm-up inference and insert a fixed cool-down interval.

Algorithm~\ref{alg:PTME} summarizes the PTME profiling procedure. The algorithm measures time, memory, and energy during inference execution and evaluates accuracy outside the measurement window. The resulting per-instance values are then analyzed statistically to obtain the reported metrics.
\begin{algorithm}[t]
\caption{PTME Profiling Procedure}
\label{alg:PTME}
\begin{algorithmic}[1]
\Require Model $m$, benchmark $D$, resource envelope $e$, decoding config $\theta$
\Ensure Per-instance result vectors of precision $\mathcal{P}$, time $\mathcal{T}$, memory $\mathcal{M}$, energy $\mathcal{E}$
\State Initialize result vectors $\mathcal{P}, \mathcal{T}, \mathcal{M}, \mathcal{E}$, $\mathcal{O}$ \Comment{ $\mathcal{O}$ stores model inference results and ground truth}
\State Select fixed benchmark subset $S \subset D$
\State Apply resource envelope $e$ \Comment{CPU binding, frequency cap, memory limit}
\State \textbf{Warm-up:} execute one inference and discard the result
\For{each instance $(p, y) \in S$}
  \State Wait fixed cool-down interval
  \State Read initial energy counter $E_0$ and start time $t_0$
  \State Execute inference with model $m$ on prompt $p$ using the decoding configuration $\theta$ and obtain output $o$
  \State Read final energy counter $E_1$ and end time $t_1$
  \State $T \gets t_1 - t_0$
  \State $E \gets E_1 - E_0$
  \State Measure peak memory usage $M$ during inference
  \State Append $T, M, E, (o, y)$ to $\mathcal{T}, \mathcal{M}, \mathcal{E}, \mathcal{O}$
\EndFor
\For{each output $(o, y) \in \mathcal{O}$}
    \State $P \gets \text{Precision}(o, y)$
    \State Append $P$ to $\mathcal{P}$
\EndFor
\State \Return $\mathcal{P}, \mathcal{T}, \mathcal{M}, \mathcal{E}$
\end{algorithmic}
\end{algorithm}

All models are executed through the same inference backend to ensure consistent invocation overhead, as described in Algorithm~\ref{alg:llmcall}. Inference calls are blocking and return only after the full response has been generated.
\begin{algorithm}[t]
\caption{Local LLM Inference Call Procedure}
\label{alg:llmcall}
\begin{algorithmic}[1]
\Require Model $m$, prompt $p$, max output tokens $L$
\Ensure Generated output $o$
\State Invoke local backend with model $m$ and prompt $p$
\State Set deterministic greedy decoding (temperature $= 0$)
\State Set maximum output tokens to $L$
\State Execute inference and wait for completion
\State \Return generated output $o$
\end{algorithmic}
\end{algorithm}

After inference, precision is evaluated in a dataset-specific manner, as summarized in Algorithm~\ref{alg:accuracy}, strictly outside the resource measurement window to avoid interference with the time, memory, and energy measurements.
\begin{algorithm}[t]
\caption{Dataset-Specific Precision Evaluation}
\label{alg:accuracy}
\begin{algorithmic}[1]
\Require Output $o$, reference answer $y$, dataset $D$
\Ensure Precision score $P \in \{0,1\}$
\If{$D$ is HumanEval}
  \State Execute the generated function in $o$ against the hidden test cases
  \State $P \gets 1$ if all tests pass (pass@1); else $0$
\ElsIf{$D$ is GSM8K}
  \State Extract the final numerical answer from $o$
  \State $P \gets 1$ if the extracted answer matches the reference answer $y$; else $0$
\ElsIf{$D$ is MMLU-Pro}
  \State Extract the predicted multiple-choice option from $o$
  \State $P \gets 1$ if the predicted option matches the reference option $y$; else $0$
\EndIf
\State \Return $P$
\end{algorithmic}
\end{algorithm}

\subsection{Statistical Analysis}
All measurements are obtained on a fixed set of benchmark instances, evaluated identically across all models and resource envelopes, yielding repeated measures in which each instance forms a matched block across conditions. We therefore use paired, non-parametric tests that respect this block structure and make no distributional assumptions.

Precision is a binary per-instance outcome (correct or incorrect). For multiple comparisons across conditions, we use Cochran's Q test; for post hoc pairwise comparisons, we use McNemar's test, both of which are appropriate for matched binary data. The remaining metrics (time, memory, energy) are continuous; for these, we use the Friedman test for multiple comparisons and Nemenyi's test for post-hoc pairwise comparisons.
When a multiple comparison is significant, we conduct post-hoc comparisons with a 5\% significance level.

\section{Experimental Design}
\label{sec:design}

This section describes the experimental design adopted to answer the research questions introduced in Section~\ref{sec:introduction}. The design is guided by three principles. First, all conclusions must be supported by \emph{measured} quantities rather than theoretical or proxy descriptors alone. Second, accuracy is treated as a first-class dimension and is never decoupled from resource efficiency. Third, the design favors minimality: only the experiments strictly necessary for each research question are included, avoiding redundant or confounding evaluations. These principles ensure both scientific rigor and reproducibility.

\subsection{Execution Environment}
\label{sec:execution_environment}

All experiments were conducted on a desktop-class machine running Ubuntu 24.04, equipped with an Intel Core i7-12700K processor and \SI{64}{\giga\byte} of DRAM. The processor has a hybrid architecture with 8 performance cores (P-cores; base \SI{3.6}{\giga\hertz}, up to \SI{4.9}{\giga\hertz} turbo) and 4 efficiency cores (E-cores), for 20 threads in total. To avoid confounding effects from heterogeneous core types, all inference is restricted to the P-cores, with CPU frequency capped per envelope as described in Section~\ref{sec:envelope_design}. All inference is executed locally through Ollama v0.23.2 as a uniform inference backend, ensuring consistent invocation overhead across models. The software environment is fixed within a Docker container, while hardware resource constraints are imposed at runtime. To reflect the resource-constrained deployment scenarios represented by the evaluated resource envelopes, all experiments are executed in CPU-only mode; no discrete or integrated GPU acceleration is used.

\begin{table*}[t]
\centering
\caption{Representative resource envelopes of edge and mobile device classes.}
\label{tab:edge_devices}
\begin{tabular}{lcccc}
\toprule
{Device Class} &
{CPU Cores} &
{Typical Frequency (GHz)} &
{Typical RAM (GB)} &
{Approx. Power (W)} \\
\midrule
Single-board computer (low-end)   & 4 & 1.2--1.5 & 4   & 5--7 \\
Single-board computer (high-end)  & 4 & 2.0--2.4 & 8   & 8--10 \\
Smartphone (mid-range)            & 6 & 1.8--2.4 & 8--12 & 5--10 \\
Smartphone (high-end)             & 8 & 2.5--3.0 & 12--16 & 10--15 \\
\bottomrule
\end{tabular}
\end{table*}
\begin{table*}[t]
\centering
\caption{Constrained desktop configurations used to approximate edge-class resource envelopes.}
\label{tab:desktop_mapping}
\begin{tabular}{lccccc}
\toprule
{Envelope} &
{P-cores} &
{Logical CPU IDs} &
{Ollama Threads} &
{CPU Freq Cap (\si{\giga\hertz})} &
{RAM Limit (\si{\giga\byte})} \\
\midrule
E1 (Tiny edge device)  & 1 & 2--3    & 1 & 1.5 & 5  \\
E2 (SBC-like device)           & 2 & 2--5    & 2 & 1.8 & 8  \\
E3 (Mid-range mobile)        & 4 & 2--9    & 4 & 2.0 & 12 \\
E4 (High-end mobile)     & 6 & 2--13   & 6 & 2.2 & 14 \\
E5 (Laptop-class device) & 8 & 0--15   & 8 & 2.5 & 16 \\
\bottomrule
\end{tabular}
\end{table*}
\begin{table*}[t]
\centering
\caption{Linux commands used to enforce constrained execution (example shown for Envelope E5).}
\label{tab:linux_constraints}
\begin{tabular}{>{\ttfamily}p{0.55\linewidth} p{0.40\linewidth}}
\toprule
{Command} & {Purpose} \\
\midrule
cpupower -c 0-15 frequency-set -g performance      & Disable dynamic frequency scaling \\
cpupower -c 0-15 frequency-set -d 2.5GHz -u 2.5GHz            & Cap minimum and maximum CPU frequency \\
docker run --rm \textbackslash &\\
\quad --cpuset-cpus="0-15" \textbackslash& Restrict inference to selected CPU cores\\
\quad --memory="16g" \textbackslash&Enforce maximum RAM usage\\
\quad --memory-swap="16g" \textbackslash&Disable swap fallback\\
\bottomrule
\end{tabular}
\end{table*}
\subsection{Resource Envelope Design}
\label{sec:envelope_design}

All experiments are executed under controlled resource constraints rather than on actual edge or mobile hardware. Our objective is to study the behavior of lightweight LLM inference under resource envelopes that are representative of edge/mobile environments, such as limited compute capacity, capped frequency, and restricted memory. Prior work on edge computing highlights the strong influence of CPU frequency, memory capacity, and parallelism on both the performance and energy characteristics of inference workloads, demonstrating that resource envelopes shape trade-offs among latency, memory usage, and energy consumption~\cite{varghese2021survey, ordonez2025intelligent}.

Table~\ref{tab:edge_devices} summarizes indicative resource envelopes of common edge and mobile device classes, based on vendor specifications and surveys of edge computing use cases. These envelopes motivate the constrained desktop configurations used in this study, but are not exact hardware targets.

Based on these envelopes, Table~\ref{tab:desktop_mapping} reports the constrained desktop configurations used to approximate similar resource availability. The listed logical CPU IDs correspond to P-cores and inference is restricted to the listed core ranges to control parallelism. For each envelope, the number of Ollama inference threads was set equal to the number of allocated P-cores, and frequency caps and memory limits were applied to approximate reduced compute and memory budgets. In the E1 envelope, the RAM limit is set to \SI{5}{\giga\byte} to ensure that the largest model, Mistral-7B, which requires approximately \SI{4.8}{\giga\byte}, can be loaded.

Table~\ref{tab:linux_constraints} summarizes the commands used to enforce the resource constraints in the Docker-based experimental environment. CPU binding and memory limits were applied to the Docker container using Docker options such as \texttt{--cpuset-cpus}, \texttt{--memory}, and \texttt{--memory-swap}, while CPU frequency caps were applied on the host system using Linux \texttt{cpupower}. These settings allow the software environment to be fixed by Docker while controlling the hardware resource constraints at runtime.

This constrained execution setup does not attempt to emulate the complete hardware stack of specific edge devices, but it reproduces key resource limitations known to affect performance and energy characteristics in mobile and edge contexts \cite{varghese2021survey, ordonez2025intelligent}. Constraining CPU parallelism, frequency, and memory availability on a desktop platform enables systematic analysis of PTME metrics under progressively tighter resource budgets, facilitating insight into trade-offs relevant for edge deployment scenarios.

\subsection{Benchmark Sampling}

Running all benchmark instances for all combinations of models and resource envelopes is computationally prohibitive in our local LLM evaluation setting. A preliminary estimate indicated that evaluating all instances across the considered benchmarks would require approximately 200 days in total. We therefore adopt a partial evaluation protocol in which a fixed subset of instances, selected randomly in advance, is used consistently across all configurations to ensure comparability.

For HumanEval, we evaluate all 164 instances because the benchmark size is relatively small. For GSM8K, we randomly sample 200 instances from the 1319 instances in the test set. For MMLU-Pro, we randomly sample 20 instances from each of the 14 categories, resulting in 280 instances out of 12032 in total. This sampling preserves coverage across all MMLU-Pro categories. Overall, the protocol substantially reduces the computational cost, lowering the estimated total evaluation time from approximately 200 days to about 7 days.

This design is justified by prior work on efficient LLM benchmarking, which showed that full-benchmark performance can be reliably estimated from substantially smaller subsets. For example, MMLU performance over approximately 14k examples was estimated from 100 carefully selected examples with an average error of about 2\%~\cite{Polo2024tinyBenchmarks}. Since the objective of this study is not to provide a fine-grained performance ranking of LLMs, but to analyze how resource constraints affect inference time, memory usage, and energy consumption, randomly sampled subsets are sufficient to make the local evaluation computationally feasible.

\subsection{Proxy Descriptors}

\begin{table}[t]
\centering
\caption{Evaluated LLMs and proxy descriptors}
\label{tab:model_info}
{
\setlength{\tabcolsep}{3pt}
\begin{tabular}{llrrrr}
\toprule
{Model} &
{Abbrev.} &
{Params (B)} &
{Loaded Mem. (\si{\giga\byte})} &
{FLOPs (T)} \\
\midrule
TinyLlama-1.1B & TL & 1.100 & 0.645 & 0.265 \\
Qwen2.5-1.5B    & QW  & 1.544 & 1.098 & 0.395 \\
Gemma-2B & GM     & 2.506 & 1.748 & 0.642 \\
Phi-2 (2.7B)& P2      & 2.780 & 2.198 & 0.678 \\
Phi-3-mini   & P3     & 3.821 & 3.731 & 0.953 \\
Mistral-7B & MS    & 7.248 & 4.833 & 1.820 \\
\bottomrule
\end{tabular}
}
\end{table}
\begin{table*}[!t]
  \centering
  \caption{Measured PTME metrics under envelope E1 (\textbf{bold}: best value; \underline{underline}: not significantly different from the best)}
  \label{tab:ptme_e1}
{
\setlength{\tabcolsep}{3pt}
\begin{tabular}{llrrrrrrrrrr}
\toprule
Bench. & Model & $P$ (\%) $\uparrow$ & $T$ (s) $\downarrow$ & $M$ (GB) $\downarrow$ & $E$ (J) $\downarrow$ & $T$/tok (ms/tok) $\downarrow$ & $E$/tok (J/tok) $\downarrow$ & TTFT (s) $\downarrow$ & TTFT/tok $\downarrow$ & In (tok) & Out (tok) \\
\midrule
\multirow{6}{*}{HumanEval}
&TL & 2.4 & 32.99 & \textbf{0.78} & 694.38 & \textbf{63.19} & \textbf{1.32} & \textbf{8.26} & \textbf{37.16} & 215 & 295 \\
&GM & 19.5 & 42.20 & 1.92 & 961.80 & 120.98 & 2.73 & 14.37 & 76.03 & 185 & 152 \\
&QW & \textbf{48.8} & \textbf{25.81} & 1.28 & \textbf{520.06} & 74.65 & 1.51 & 9.72 & 52.34 & 179 & 159 \\
&P2 & 34.8 & 74.77 & 2.25 & 1623.43 & 167.31 & 3.63 & 22.28 & 101.87 & 206 & 216 \\
&P3 & 40.2 & 69.15 & 3.65 & 1453.95 & 202.51 & 4.24 & 27.50 & 147.61 & 182 & 148 \\
&MS & \underline{42.7} & 131.49 & 4.76 & 3372.01 & 344.25 & 8.83 & 50.23 & 275.52 & 178 & 200 \\
\midrule
\multirow{6}{*}{GSM8K}
&TL & 2.5 & \textbf{17.74} & \textbf{0.77} & \textbf{498.67} & \textbf{54.13} & \textbf{1.53} & \textbf{3.77} & \textbf{28.07} & 132 & 179 \\
&GM & 9.5 & \underline{18.48} & 1.89 & \underline{518.65} & 103.18 & 2.90 & 6.95 & 63.13 & 109 & 64 \\
&QW & 53.5 & \underline{21.20} & 1.23 & \underline{674.59} & 65.72 & 2.08 & 4.39 & 37.67 & 114 & 182 \\
&P2 & 24.0 & \underline{28.02} & 2.23 & \underline{833.50} & 107.81 & 3.21 & 8.33 & 66.42 & 123 & 91 \\
&P3 & \textbf{66.5} & 72.86 & 3.63 & 2446.97 & 209.38 & 7.02 & 11.94 & 112.71 & 105 & 231 \\
&MS & 29.0 & 106.75 & 4.72 & 3483.38 & 322.49 & 10.52 & 21.63 & 216.26 & 98 & 220 \\
\midrule
\multirow{6}{*}{MMLU-Pro}
&TL & 9.6 & \textbf{34.94} & \textbf{0.79} & \textbf{822.87} & \textbf{59.89} & \textbf{1.41} & \textbf{12.57} & \textbf{38.23} & 308 & 248 \\
&GM & 13.9 & 55.09 & 1.96 & 1043.44 & 131.81 & 2.50 & 23.69 & 86.00 & 266 & 145 \\
&QW & \underline{28.2} & 62.16 & 1.31 & 1187.12 & 96.00 & 1.83 & 17.40 & 61.33 & 268 & 365 \\
&P2 & 25.4 & 73.95 & 2.30 & 1402.81 & 167.13 & 3.17 & 39.04 & 124.19 & 286 & 114 \\
&P3 & \underline{34.3} & 170.33 & 3.69 & 4433.95 & 241.20 & 6.29 & 42.24 & 143.43 & 279 & 418 \\
&MS & \textbf{34.6} & 208.76 & 4.79 & 4315.83 & 390.96 & 7.97 & 86.27 & 309.40 & 269 & 251 \\
\bottomrule
\end{tabular}

}
\end{table*}
\begin{table*}[!t]
  \centering
  \caption{Measured PTME metrics under envelope E5 (\textbf{bold}: best value; \underline{underline}: not significantly different from the best)}
  \label{tab:ptme_e5}
{
\setlength{\tabcolsep}{3pt}
\begin{tabular}{llrrrrrrrrrr}
\toprule
Bench. & Model & $P$ (\%) $\uparrow$ & $T$ (s) $\downarrow$ & $M$ (GB) $\downarrow$ & $E$ (J) $\downarrow$ & $T$/tok (ms/tok) $\downarrow$ & $E$/tok (J/tok) $\downarrow$ & TTFT (s) $\downarrow$ & TTFT/tok $\downarrow$ & In (tok) & Out (tok) \\
\midrule
\multirow{6}{*}{HumanEval}
&TL & 2.4 & \underline{5.82} & \textbf{0.97} & 250.38 & \textbf{11.16} & \textbf{0.48} & 1.68 & 7.60 & 215 & 295 \\
&GM & 19.5 & 7.92 & 2.18 & 331.72 & 22.90 & 0.95 & 3.09 & 16.46 & 185 & 152 \\
&QW & \textbf{48.8} & \textbf{4.71} & 1.50 & \textbf{200.88} & 13.67 & 0.58 & \textbf{1.30} & \textbf{7.13} & 179 & 159 \\
&P2 & 34.8 & 13.26 & 2.47 & 556.46 & 29.88 & 1.25 & 4.31 & 19.99 & 206 & 216 \\
&P3 & 40.2 & 11.99 & 3.86 & 511.66 & 35.35 & 1.50 & 5.57 & 29.96 & 182 & 148 \\
&MS & \underline{42.7} & 21.39 & 4.95 & 985.32 & 56.16 & 2.59 & 6.46 & 35.44 & 178 & 200 \\
\midrule
\multirow{6}{*}{GSM8K}
&TL & 2.5 & \textbf{3.26} & \textbf{0.84} & \textbf{144.80} & \textbf{10.07} & \textbf{0.44} & 0.84 & \underline{6.31} & 132 & 179 \\
&GM & 9.5 & \underline{3.69} & 2.04 & \underline{159.27} & 20.81 & 0.89 & 1.57 & 14.32 & 109 & 64 \\
&QW & 53.5 & 4.61 & 1.40 & \underline{198.87} & 14.00 & 0.60 & \textbf{0.72} & \textbf{6.25} & 114 & 182 \\
&P2 & 24.0 & \underline{5.27} & 2.34 & \underline{225.81} & 20.85 & 0.88 & 1.78 & 14.22 & 123 & 91 \\
&P3 & \textbf{66.5} & 12.56 & 3.71 & 575.66 & 36.29 & 1.66 & 2.66 & 25.08 & 105 & 231 \\
&MS & 29.0 & 19.49 & 4.77 & 912.99 & 58.09 & 2.73 & 2.95 & 29.43 & 98 & 220 \\
\midrule
\multirow{6}{*}{MMLU-Pro}
&TL & 9.6 & \textbf{4.80} & \textbf{0.81} & \textbf{233.72} & \textbf{8.26} & \textbf{0.40} & \textbf{1.41} & \textbf{4.35} & 308 & 248 \\
&GM & 13.9 & 8.87 & 2.11 & 371.41 & 21.22 & 0.89 & 4.19 & 15.23 & 266 & 145 \\
&QW & \underline{28.2} & 10.75 & 1.42 & 496.60 & 16.72 & 0.77 & 2.05 & 7.35 & 268 & 365 \\
&P2 & 25.4 & 11.39 & 2.32 & 498.43 & 26.14 & 1.14 & 6.11 & 19.82 & 286 & 114 \\
&P3 & \underline{34.3} & 28.78 & 3.73 & 1294.36 & 40.73 & 1.83 & 8.08 & 28.08 & 279 & 418 \\
&MS & \textbf{34.6} & 28.54 & 4.82 & 1322.23 & 53.02 & 2.46 & 9.35 & 33.17 & 269 & 251 \\
\bottomrule
\end{tabular}

}
\end{table*}
Table~\ref{tab:model_info} summarizes the model-scale proxy descriptors used in our analysis for the six evaluated LLMs with their abbreviations: parameter count, loaded model memory, and FLOPs. Loaded model memory is the memory required to load each model in the Ollama environment. FLOPs are estimated using THOP~\cite{thop} with a dummy input of shape $[1,128]$, corresponding to a single forward pass with a batch size of 1 and a sequence length of 128 tokens; they thus provide an architecture-level estimate of per-pass computational cost rather than the total cost of a full inference. All three are static quantities known prior to execution and can therefore serve as indicators for model selection before inference.

\section{Results and Discussion}
\label{sec:results}

This section presents and discusses the experimental results. We first characterize the basic behavior of the measured PTME metrics, and then organize the analysis according to the three research questions.

\subsection{Preliminaries}
\label{subsec:preliminaries}

Before addressing the research questions, this section characterizes the basic behavior of the PTME metrics. Tables~\ref{tab:ptme_e1} and~\ref{tab:ptme_e5} report the PTME metrics for all evaluated models under the most constrained envelope (E1) and the least constrained envelope (E5), respectively. The results for all envelopes are reported in the Supplementary material. For each model, we report means of precision $P$, time $T$, memory $M$, and energy $E$, together with the time to first token (TTFT) and the input and output token counts. To enable fair comparison across models and benchmarks with different sequence lengths, we additionally report token-normalized metrics. $T$ and $E$ are normalized by the total number of tokens (input plus output), since they reflect the cost of the entire inference. In contrast, TTFT is normalized by the number of input tokens, since it reflects the cost of processing the prompt before generation begins. In the table, the best value (maximum in precision and minimum in the others) is represented in bold, while values that are not significantly different from the best are underlined.

From Tables~\ref{tab:ptme_e1} and~\ref{tab:ptme_e5}, the relative ordering of models is largely consistent across both envelopes. The multiple-comparison tests confirm significant differences among models for all PTME metrics across all benchmarks ($p<0.01$), indicating that model choice significantly affects not only precision but also all cost metrics. 

In terms of precision, the best-performing model differs by benchmark. The relatively modest precision values across all evaluated models indicate that the selected benchmark instances remain challenging even for current lightweight LLMs. This characteristic facilitates meaningful comparison of PTME trade-offs by avoiding saturation effects in model performance.
For all other metrics, the smallest model (TinyLlama) is generally the most efficient, as expected from its size. The exceptions are the absolute time and energy on HumanEval and the TTFT on GSM8K, where Qwen2.5 is the best. This is because Qwen2.5 produces fewer tokens on HumanEval and GSM8K, which reduces these non-normalized quantities despite its larger size. On GSM8K, energy and time of Gemma, Qwen2.5, and Phi-2 are not significantly different from those of TinyLlama. This is because the input and output token sizes of these models are smaller than those of TinyLlama, thereby reducing total time and energy.

\begin{figure*}[!tb]
    \centering
    \begin{minipage}[t]{0.32\textwidth}
        \includegraphics[width=\linewidth]{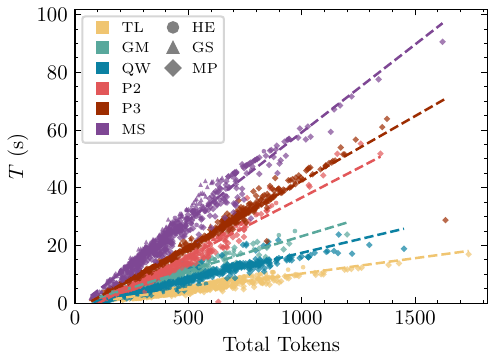}
        \subcaption{Tokens vs. Time}
        \label{fig:tokens_time}
    \end{minipage}\hfill
    \begin{minipage}[t]{0.32\textwidth}
        \includegraphics[width=\linewidth]{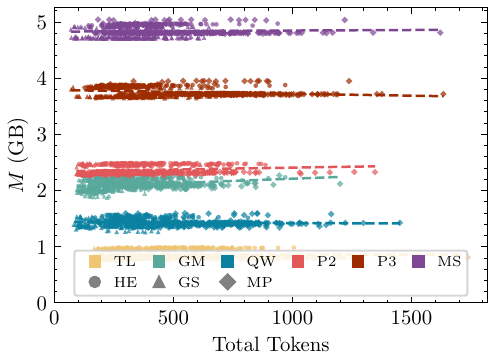}
        \subcaption{Tokens vs. Memory}
        \label{fig:tokens_memory}
    \end{minipage}\hfill
    \begin{minipage}[t]{0.32\textwidth}
        \includegraphics[width=\linewidth]{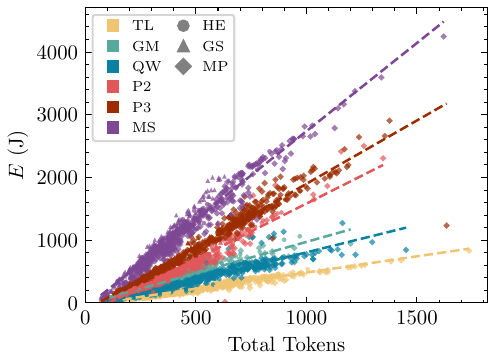}
        \subcaption{Tokens vs. Energy}
        \label{fig:tokens_energy}
    \end{minipage}
    \caption{Relationship between total tokens and time, memory, and energy metrics on envelope E5 across all benchmarks}
    \label{fig:tokens_e5}
\end{figure*}
\begin{figure*}[tb]
    \centering
    \begin{minipage}[t]{0.24\textwidth}
        \includegraphics[width=\linewidth]{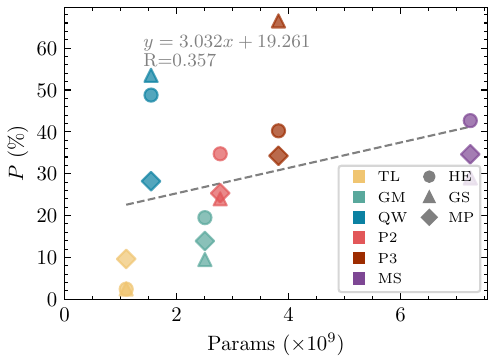}
        \subcaption{Params. vs. Precision}
        \label{fig:param_prec_HE}
    \end{minipage}\hfill
    \begin{minipage}[t]{0.24\textwidth}
        \includegraphics[width=\linewidth]{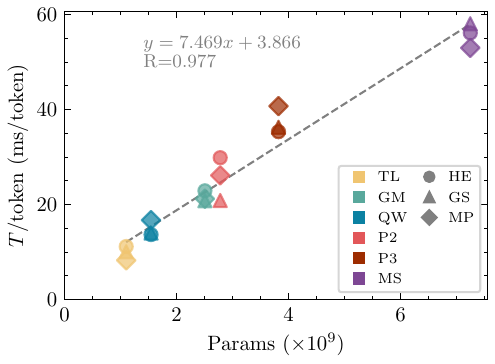}
        \subcaption{Params. vs. Time/token}
        \label{fig:param_time_HE}
    \end{minipage}\hfill
    \begin{minipage}[t]{0.24\textwidth}
        \includegraphics[width=\linewidth]{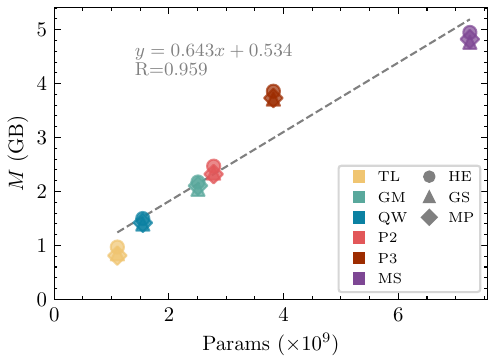}
        \subcaption{Params. vs. Memory usage}
        \label{fig:param_memory_HE}
    \end{minipage}\hfill
    \begin{minipage}[t]{0.24\textwidth}
        \includegraphics[width=\linewidth]{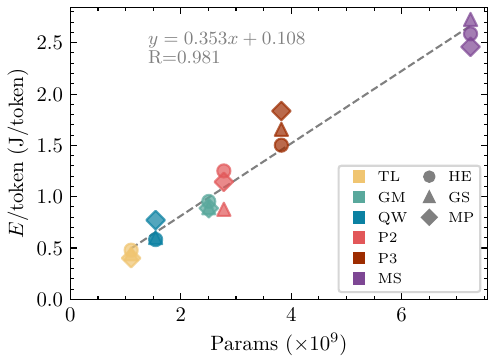}
        \subcaption{Params.  vs. Energy/tokens}
        \label{fig:param_energy_HE}
    \end{minipage}
    \centering
    \begin{minipage}[t]{0.24\textwidth}
        \includegraphics[width=\linewidth]{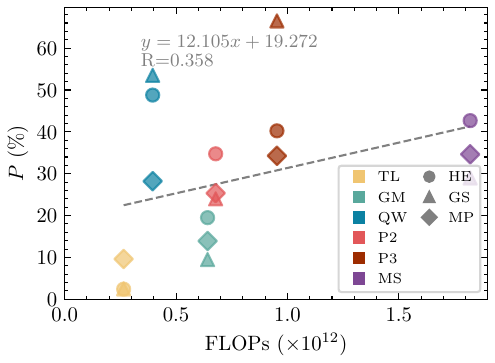}
        \subcaption{FLOPs vs. Precision}
        \label{fig:flops_prec_HE}
    \end{minipage}\hfill
    \begin{minipage}[t]{0.24\textwidth}
        \includegraphics[width=\linewidth]{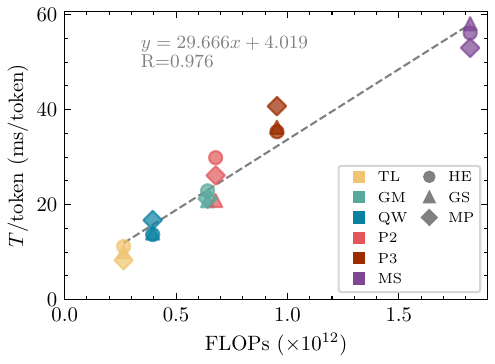}
        \subcaption{FLOPs vs. Time/token}
        \label{fig:flops_time_HE}
    \end{minipage}\hfill
    \begin{minipage}[t]{0.24\textwidth}
        \includegraphics[width=\linewidth]{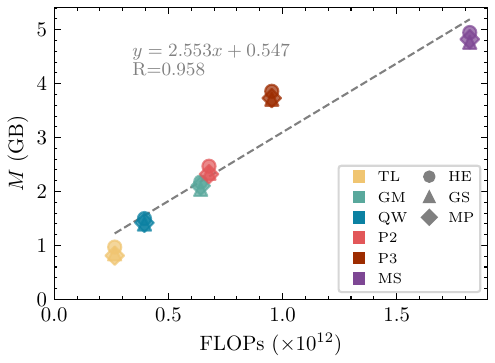}
        \subcaption{FLOPs vs. Memory usage}
        \label{fig:flops_memory_HE}
    \end{minipage}\hfill
    \begin{minipage}[t]{0.24\textwidth}
        \includegraphics[width=\linewidth]{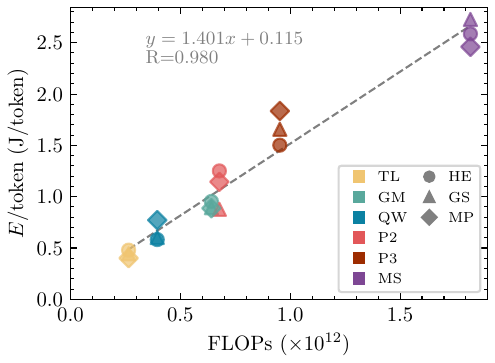}
        \subcaption{FLOPs vs. Energy/token}
        \label{fig:flops_energy_HE}
    \end{minipage}
    \centering
    \begin{minipage}[t]{0.24\textwidth}
        \includegraphics[width=\linewidth]{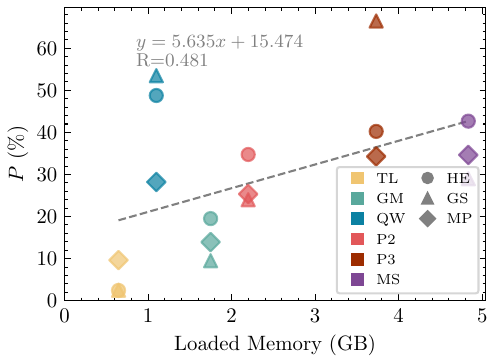}
        \subcaption{Model size vs. Precision}
        \label{fig:size_prec_HE}
    \end{minipage}\hfill
    \begin{minipage}[t]{0.24\textwidth}
        \includegraphics[width=\linewidth]{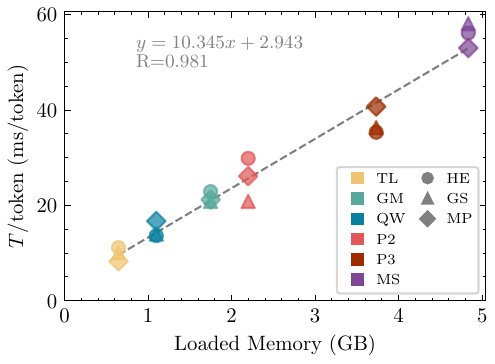}
        \subcaption{Model size vs. Time/token}
        \label{fig:size_time_HE}
    \end{minipage}\hfill
    \begin{minipage}[t]{0.24\textwidth}
        \includegraphics[width=\linewidth]{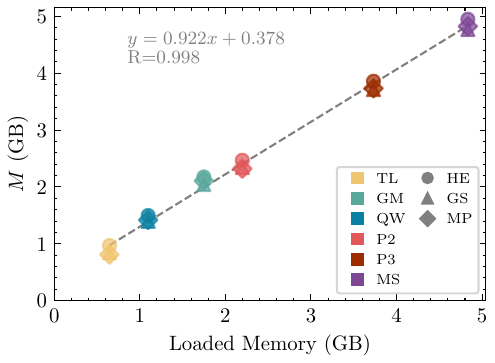}
        \subcaption{Model size vs. Memory usage}
        \label{fig:size_memory_HE}
    \end{minipage}\hfill
    \begin{minipage}[t]{0.24\textwidth}
        \includegraphics[width=\linewidth]{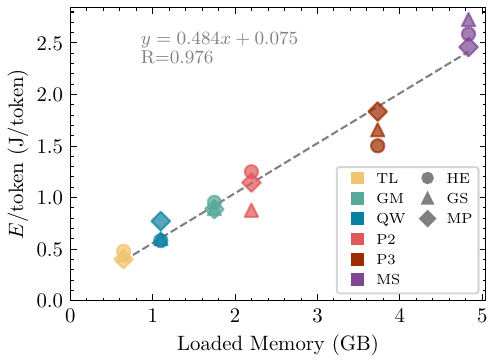}
        \subcaption{Model size vs. Energy/token}
        \label{fig:size_energy_HE}
    \end{minipage}
    \centering
    \caption{Relationship between proxies and precision, time, memory, and energy on envelope E5 across all benchmarks}
    \label{fig:proxy_e5}
\end{figure*}
We next analyze how inference cost varies with the length of the input and output sequences. Fig.~\ref{fig:tokens_e5} shows the relationship between the total token count and time, memory, and energy under envelope E5. The horizontal axis shows the total tokens, while the vertical axis shows each metric. 

From Fig.~\ref{fig:tokens_e5}, time and energy increase approximately linearly with the total tokens, whereas memory is almost constant regardless of the token count. This indicates that time and energy are governed by the number of tokens, while memory is determined by the model footprint rather than the sequence length. The same trend was observed for the other envelopes (Supplementary). Consequently, the effect of sequence length can be controlled through token normalization for time and energy, whereas for memory it need not be considered. 

\subsection{RQ1: Proxy Descriptors versus Measured PTME Metrics}
\label{sec:rq1_design}
RQ1 examines the extent to which static model-scale descriptors (proxies) correlate with the measured PTME metrics. We consider three proxies: parameter count, theoretical FLOPs, and loaded memory (Table~\ref{tab:model_info}). 

Fig.~\ref{fig:proxy_e5} shows the relationship between each proxy and the measured PTME metrics under envelope E5, together with the fitted regression lines and Pearson correlation coefficients $R$. The cost metrics ($T$/token, $M$, $E$/token) all exhibit strong linear correlations with the proxies ($R\approx0.95$--1.0). In particular, measured memory usage is almost perfectly correlated with loaded memory ($R\approx0.998$--1.0), confirming that runtime memory consumption is largely determined by the loaded model footprint. 
Precision, in contrast, scatters widely around its regression lines, with $R$ below 0.5 for every proxy. Thus, a clear contrast---strong correlation with the cost metrics but no correlation with precision---emerges within a single envelope.

\begin{table}[tb]
\centering
\caption{Pearson correlation coefficient between proxy descriptors and measured metrics}
\label{tab:correlation}
{
\setlength{\tabcolsep}{3pt}
\begin{tabular}{llrrrr}
\toprule
&& $P$ (\%) & $T$/tok (ms/tok) & $M$ (GB) & $E$/tok (J/tok) \\
\midrule
\multirow{3}{*}{E1}
&Params & 0.357 & \textbf{0.975} & 0.961 & 0.956 \\
&FLOPs & 0.358 & 0.974 & 0.960 & 0.955 \\
&Loaded Memory & \textbf{0.481} & 0.964 & \textbf{0.999} & \textbf{0.957} \\
\midrule
\multirow{3}{*}{E2}
&Params & 0.357 & \textbf{0.973} & 0.959 & 0.946 \\
&FLOPs & 0.358 & 0.972 & 0.958 & 0.944 \\
&Loaded Memory & \textbf{0.481} & 0.971 & \textbf{0.998} & \textbf{0.970} \\
\midrule
\multirow{3}{*}{E3}
&Params & 0.357 & 0.947 & 0.958 & \textbf{0.980} \\
&FLOPs & 0.358 & 0.945 & 0.957 & 0.979 \\
&Loaded Memory & \textbf{0.481} & \textbf{0.957} & \textbf{0.998} & 0.971 \\
\midrule
\multirow{3}{*}{E4}
&Params & 0.357 & 0.958 & 0.959 & 0.971 \\
&FLOPs & 0.358 & 0.957 & 0.958 & 0.970 \\
&Loaded Memory & \textbf{0.481} & \textbf{0.966} & \textbf{0.998} & \textbf{0.974} \\
\midrule
\multirow{3}{*}{E5}
&Params & 0.357 & 0.977 & 0.959 & \textbf{0.981} \\
&FLOPs & 0.358 & 0.976 & 0.958 & 0.980 \\
&Loaded Memory & \textbf{0.481} & \textbf{0.981} & \textbf{0.998} & 0.976 \\
\bottomrule
\end{tabular}

}
\end{table}
This contrast is consistent across all envelopes. Table~\ref{tab:correlation} reports the Pearson correlation coefficients between the proxies and the PTME metrics under envelopes E1--E5. Precision, in contrast, shows only a weak correlation with each proxy across all envelopes ($R \approx 0.36$ for parameter count and FLOPs, $R \approx 0.48$ for loaded memory), in sharp contrast to the strong correlations observed for the cost metrics. The rank correlation is slightly higher (Spearman $\rho \approx 0.52$, see Supplementary) but still indicates that scale provides no reliable basis for predicting which model is more accurate. Static descriptors, therefore, cannot account for task precision. This is the central finding of RQ1: proxies approximate inference cost well but fail to predict precision. Among the proxies, although the strongest proxy for time per token and energy per token differs by envelope, loaded memory is the most informative. It almost perfectly explains measured memory ($R\approx0.998$--$1.0$ across all envelopes) and better accounts for the cost of Phi-3 than the other proxies, since Phi-3 has a relatively large memory footprint given its parameter count and FLOPs.

In summary, static proxy descriptors are effective as first-order approximations of inference cost---particularly token-normalized time, memory, and energy---but are insufficient for predicting precision. Proxies can thus support preliminary model selection, but direct PTME measurement is indispensable whenever precision matters. This motivates RQ2, which examines the interaction between measured precision and resource consumption.

\subsection{RQ2: Precision-Cost Trade-offs and Non-Dominated Configurations}
\label{sec:rq2_design}
\begin{figure*}[tb]
  \centering
    \begin{minipage}[t]{0.32\linewidth}
      \includegraphics[width=\linewidth]{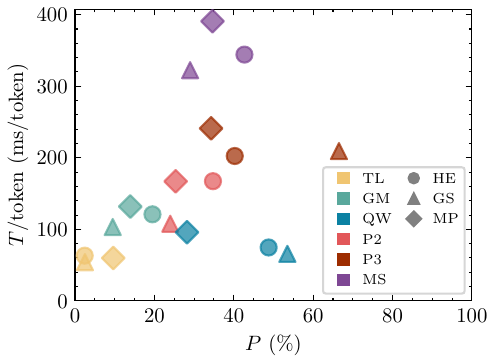}
      \subcaption{Precision vs. Time per token}
      \label{fig:perf_time_per_token_e1}
    \end{minipage}\hfill
    \begin{minipage}[t]{0.32\linewidth}
      \includegraphics[width=\linewidth]{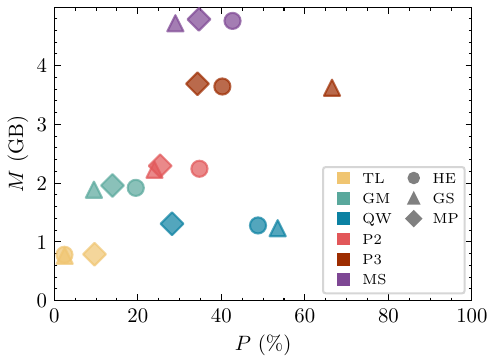}
      \subcaption{Precision vs. Memory}
      \label{fig:perf_memory_e1}
    \end{minipage}\hfill
    \begin{minipage}[t]{0.32\linewidth}
      \includegraphics[width=\linewidth]{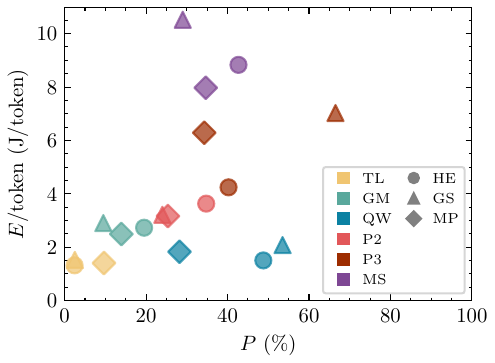}
      \subcaption{Precision vs. Energy per token}
      \label{fig:perf_energy_per_token_e1}
    \end{minipage}
    \caption{Relation between precision and energy/token, time/token, and memory metrics under envelope E1 across all benchmarks}
    \label{fig:precision_cost_e1}
    \vspace{1em}
  \centering
    \begin{minipage}[t]{0.32\linewidth}
      \includegraphics[width=\linewidth]{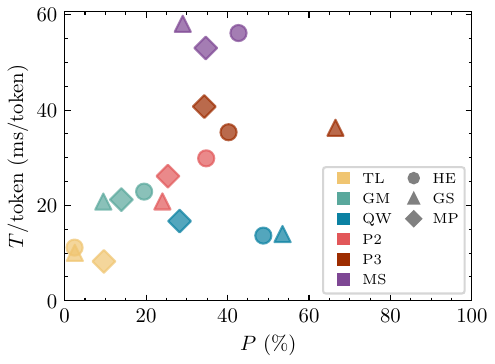}
      \subcaption{Precision vs. Time per token}
      \label{fig:perf_time_per_token_e5}
    \end{minipage}\hfill
    \begin{minipage}[t]{0.32\linewidth}
      \includegraphics[width=\linewidth]{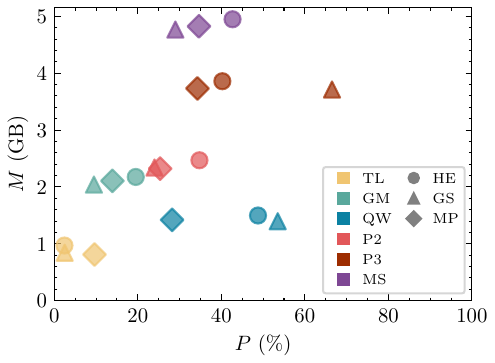}
      \subcaption{Precision vs. Memory}
      \label{fig:perf_memory_e5}
    \end{minipage}\hfill
    \begin{minipage}[t]{0.32\linewidth}
      \includegraphics[width=\linewidth]{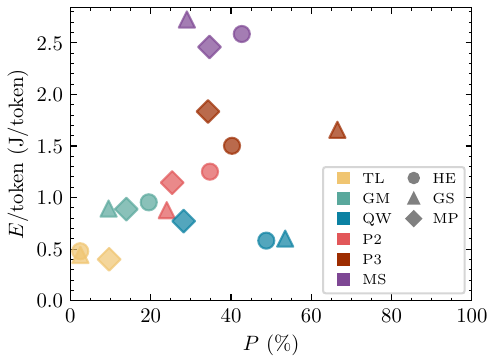}
      \subcaption{Precision vs. Energy per token}
      \label{fig:perf_energy_per_token_e5}
    \end{minipage}
    \caption{Relation between precision and energy/token, time/token, and memory metrics under envelope E5 across all benchmarks}
    \label{fig:precision_cost_e5}
\end{figure*}
RQ2 examines how precision trades off against the three cost metrics across models under a fixed resource envelope, and which configurations are non-dominated and most appropriate for practical use. We use the most constrained envelope (E1) and the least constrained envelope (E5) as representative cases; the same analysis for the other envelopes is provided in the Supplementary material.

Figs.~\ref{fig:precision_cost_e1} and \ref{fig:precision_cost_e5} show the relationship between precision and the three cost metrics (time per token, memory, energy per token), with models distinguished by color and benchmarks by shape. The horizontal axis shows the precision, while the vertical axis shows the three cost metrics. Since higher precision and lower cost are preferable, configurations toward the lower-right are more desirable. 

The lowest-cost model (TinyLlama) achieves extremely low precision across all benchmarks, showing that speed and resource frugality alone do not yield practical inference. Conversely, the largest model (Mistral) occupies the high-cost region, yet its precision gains are benchmark-dependent and not necessarily commensurate with these costs. Qwen2.5, in contrast, achieves relatively high precision on HumanEval and GSM8K while maintaining low per-token time, energy, and memory usage. Phi-3 shows a different position: the precision is relatively high, and the cost metrics are moderate. These two-metric views already reveal cases in which one model surpasses the other on both axes.

Another noteworthy observation is that the relationship between execution time and energy consumption differs between the two representative resource envelopes. Under the most constrained envelope (E1), the time and energy plots exhibit similar overall trends but are not identical, indicating that reducing execution time does not always minimize energy consumption. In contrast, under E5, the two plots become much more similar, suggesting that execution time becomes a better proxy for energy consumption in less resource-constrained environments. This result highlights the importance of directly measuring energy when selecting models for highly constrained devices.

\begin{figure}[tb]
    \centering
    \includegraphics[width=\columnwidth]{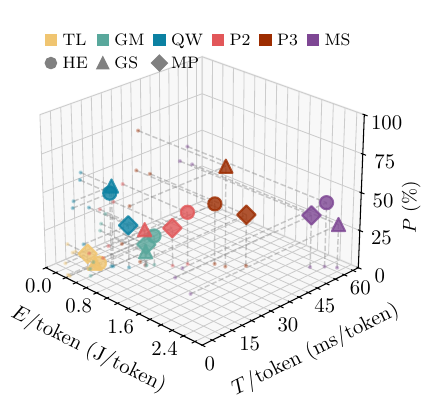}
    \caption{3D visualizations of precision, time per token, and energy per token under envelope E5 across all benchmarks}
    \label{fig:3d_e5}
\end{figure}
Fig.~\ref{fig:3d_e5} provides a three-dimensional view of precision, time per token, and energy per token under envelope E5, where each point is a model-benchmark combination. When viewed jointly across the three axes, the models fall broadly into four groups: TinyLlama and Gemma form a low-cost, low-precision group near the origin. Phi-2 and Phi-3 occupy an intermediate-cost, moderate-to-high-precision region; Mistral lies farthest from the origin, incurring the highest time and energy without a commensurate precision gain; and Qwen2.5 stands apart, remaining in the low-cost region yet achieving precision comparable to the more expensive models. This isolation of Qwen2.5---low cost together with high precision---is precisely what the three-dimensional view reveals and what the two-metric projections only partially convey.

\begin{table}[tb]
\caption{Single-metric optimal models versus non-dominated sets under envelopes E1 and E5}
\label{tab:rq4_pareto}
\centering
{
\setlength{\tabcolsep}{3.5pt}
\begin{tabular}{lllllll}
\toprule
&& \multicolumn{4}{c}{Single-metric optimum} & \\
\cmidrule(lr){3-6}
Env.&Bench. & $P$ & $T$/tok & $M$ & $E$/tok & Non-dom. set \\
\midrule
\multirow{3}{*}{E1}
&HE & QW (MS) & TL & TL & TL & TL, QW \\
&GS & P3 & TL & TL & TL & TL, QW, P3 \\
&MP & MS (QW, P3) & TL & TL & TL & TL, QW, P3, MS \\
\midrule
\multirow{3}{*}{E5}
&HE & QW (MS) & TL & TL & TL & TL, QW \\
&GS & P3 & TL & TL & TL & TL, QW, P3 \\
&MP & MS (QW, P3) & TL & TL & TL & TL, QW, P3, MS \\
\bottomrule
\end{tabular}
}
\end{table}
Table~\ref{tab:rq4_pareto} reports the non-dominated sets together with the single-metric optimum for each benchmark. In all single-metric columns, models shown in parentheses are not significantly different from the best model for the corresponding metric. In this table, such statistically comparable models appear only in the precision column, indicating that the cost-optimal model is uniquely identified for $T$/tok, $M$, and $E$/tok, whereas precision admits multiple statistically comparable choices in some benchmarks.

\begin{table*}[t]
  \centering
  \caption{Measured PTME metrics on HumanEval across all envelopes (\textbf{bold}: best value; \underline{underline}: not significantly different from the best)}
  \label{tab:ptme_humaneval}
{
\setlength{\tabcolsep}{3pt}
\begin{tabular}{llrrrrrrrrrr}
\toprule
Model & Env. & $P$ (\%) & $T$ (s) & $M$ (GB) & $E$ (J) & $T$/tok (ms/tok) & $E$/tok (J/tok) & TTFT (s) & TTFT/tok (ms/tok) & In (tok) & Out (tok) \\
\midrule
\multirow{5}{*}{TL}
&E1 & \textbf{2.4} & 32.99 & \textbf{0.78} & 694.38 & 63.19 & 1.32 & 8.26 & 37.16 & 215 & 295 \\
&E2 & \textbf{2.4} & 21.41 & 0.96 & 452.26 & 40.96 & 0.87 & 5.23 & 23.50 & 215 & 295 \\
&E3 & \textbf{2.4} & 12.74 & 0.97 & 324.51 & 24.45 & 0.62 & 3.82 & 17.21 & 215 & 295 \\
&E4 & \textbf{2.4} & 8.01 & 0.97 & 267.50 & 15.38 & 0.51 & 2.44 & 11.00 & 215 & 295 \\
&E5 & \textbf{2.4} & \textbf{5.82} & 0.97 & \textbf{250.38} & \textbf{11.16} & \textbf{0.48} & \textbf{1.68} & \textbf{7.60} & 215 & 295 \\
\midrule
\multirow{5}{*}{GM}
&E1 & \textbf{19.5} & 42.20 & \textbf{1.92} & 961.80 & 120.98 & 2.73 & 14.37 & 76.03 & 185 & 152 \\
&E2 & \textbf{19.5} & 25.00 & 2.10 & 549.23 & 71.57 & 1.57 & 8.47 & 45.02 & 185 & 152 \\
&E3 & \textbf{19.5} & 17.01 & 2.12 & 436.62 & 49.23 & 1.26 & 7.04 & 37.38 & 185 & 152 \\
&E4 & \textbf{19.5} & 11.00 & 2.13 & 366.50 & 31.84 & 1.06 & 4.50 & 23.92 & 185 & 152 \\
&E5 & \textbf{19.5} & \textbf{7.92} & 2.18 & \textbf{331.72} & \textbf{22.90} & \textbf{0.95} & \textbf{3.09} & \textbf{16.46} & 185 & 152 \\
\midrule
\multirow{5}{*}{QW}
&E1 & \textbf{48.8} & 25.81 & \textbf{1.28} & 520.06 & 74.65 & 1.51 & 9.72 & 52.34 & 179 & 159 \\
&E2 & \textbf{48.8} & 16.81 & 1.46 & 354.28 & 49.34 & 1.03 & 6.13 & 33.67 & 179 & 159 \\
&E3 & \textbf{48.8} & 9.29 & 1.48 & 247.33 & 26.96 & 0.72 & 3.27 & 17.78 & 179 & 159 \\
&E4 & \textbf{48.8} & 5.85 & 1.50 & \underline{202.61} & 16.99 & \underline{0.59} & 1.93 & 10.51 & 179 & 159 \\
&E5 & \textbf{48.8} & \textbf{4.71} & 1.50 & \textbf{200.88} & \textbf{13.67} & \textbf{0.58} & \textbf{1.30} & \textbf{7.13} & 179 & 159 \\
\midrule
\multirow{5}{*}{P2}
&E1 & \textbf{34.8} & 74.77 & \textbf{2.25} & 1623.43 & 167.31 & 3.63 & 22.28 & 101.87 & 206 & 216 \\
&E2 & \textbf{34.8} & 43.59 & 2.46 & 952.13 & 97.69 & 2.13 & 12.41 & 57.22 & 206 & 216 \\
&E3 & \textbf{34.8} & 29.02 & 2.46 & 750.11 & 65.54 & 1.69 & 9.93 & 45.79 & 206 & 216 \\
&E4 & \textbf{34.8} & 18.27 & 2.46 & 607.73 & 41.33 & 1.37 & 6.32 & 29.26 & 206 & 216 \\
&E5 & \textbf{34.8} & \textbf{13.26} & 2.47 & \textbf{556.46} & \textbf{29.88} & \textbf{1.25} & \textbf{4.31} & \textbf{19.99} & 206 & 216 \\
\midrule
\multirow{5}{*}{P3}
&E1 & \textbf{40.2} & 69.15 & \textbf{3.65} & 1453.95 & 202.51 & 4.24 & 27.50 & 147.61 & 182 & 148 \\
&E2 & \textbf{40.2} & 47.17 & 3.85 & 960.92 & 140.61 & 2.84 & 19.22 & 106.10 & 182 & 148 \\
&E3 & \textbf{40.2} & 27.41 & 3.85 & 707.22 & 80.83 & 2.08 & 12.84 & 69.21 & 182 & 148 \\
&E4 & \textbf{40.2} & 17.05 & 3.86 & 572.87 & 50.18 & 1.68 & 8.10 & 43.40 & 182 & 148 \\
&E5 & \textbf{40.2} & \textbf{11.99} & 3.86 & \textbf{511.66} & \textbf{35.35} & \textbf{1.50} & \textbf{5.57} & \textbf{29.96} & 182 & 148 \\
\midrule
\multirow{5}{*}{MS}
&E1 & \textbf{42.7} & 131.49 & \textbf{4.76} & 3372.01 & 344.25 & 8.83 & 50.23 & 275.52 & 178 & 200 \\
&E2 & \textbf{42.7} & 92.95 & 4.94 & 2048.76 & 241.70 & 5.32 & 36.14 & 195.02 & 178 & 200 \\
&E3 & \textbf{42.7} & 46.23 & 4.94 & 1263.75 & 121.05 & 3.31 & 17.50 & 95.77 & 178 & 200 \\
&E4 & \textbf{42.7} & 28.36 & 4.94 & 1022.94 & 74.37 & 2.68 & 9.88 & 54.15 & 178 & 200 \\
&E5 & \textbf{42.7} & \textbf{21.39} & 4.95 & \textbf{985.32} & \textbf{56.16} & \textbf{2.59} & \textbf{6.46} & \textbf{35.44} & 178 & 200 \\
\bottomrule
\end{tabular}

}
\end{table*}
From this table, the non-dominated sets differ by benchmark: \{TinyLlama, Qwen2.5\} for HumanEval, \{TinyLlama, Qwen2.5, Phi-3\} for GSM8K, and \{TinyLlama, Qwen2.5, Phi-3, Mistral\} for MMLU-Pro.
Comparing these sets with the single-metric optima shows that no single metric captures the full set, even when models that are not significantly different from the best are also considered. Under the precision metric alone, the best model varies by task, whereas under each of the three cost metrics, TinyLlama is selected across all benchmarks. Thus, every single metric selects only an extreme of the front---the most accurate or the cheapest model---and an efficiency-only evaluation collapses to a single model regardless of task. 
Considering all four objectives jointly reveals configurations that are not selected by any single metric but remain practically relevant. For example, Qwen2.5 is non-dominated on GSM8K despite not being the precision-optimal or cost-optimal model. More importantly, on MMLU-Pro, Qwen2.5 and Phi-3 are non-dominated and are not significantly different from Mistral in precision, while requiring substantially lower resource cost. This means that, after accounting for statistical significance in precision, Qwen2.5 and Phi-3 provide viable alternatives to Mistral that reduce time, memory, and energy without a statistically confirmed loss in precision. The non-dominated sets also show that the largest model is not necessarily preferable. On HumanEval and GSM8K, Mistral drops out of the non-dominated set even under the least-constrained envelope E5. On MMLU-Pro, Mistral re-enters the non-dominated set, but it is not the only defensible choice because Qwen2.5 and Phi-3 offer lower-cost alternatives with statistically comparable precision.

Which configuration to choose from a non-dominated set depends on the relative preference placed on precision versus cost. The Pareto analysis should therefore be interpreted as a decision-support tool, not as a procedure for selecting a single best model: it removes dominated configurations and exposes the remaining precision–cost alternatives. The most informative region is often the front knee, where precision is largely retained at substantially lower cost. For example, in MMLU-Pro, although Mistral achieves the highest mean precision, Qwen2.5 reaches 28.2\% precision at roughly one-third of the time and energy of Mistral (34.6\%), and Qwen2.5 and Phi-3 are not significantly different from Mistral in precision. Thus, the largest model buys only a limited precision advantage at a much higher physical cost. This result indicates that, especially under edge-class envelopes where latency and energy budgets are tight, lower-cost non-dominated models such as Qwen2.5 or Phi-3 can be more suitable choices than the largest model. On desktop-class deployments, larger models remain viable when precision is paramount, but the Pareto view clarifies whether their additional precision is worth the extra time, memory, and energy.

\subsection{RQ3: Sensitivity of PTME to the Resource Envelope}
\label{sec:rq3_design}
\begin{figure*}[tb]
    \centering
    \begin{minipage}[t]{0.32\textwidth}
        \includegraphics[width=\linewidth]{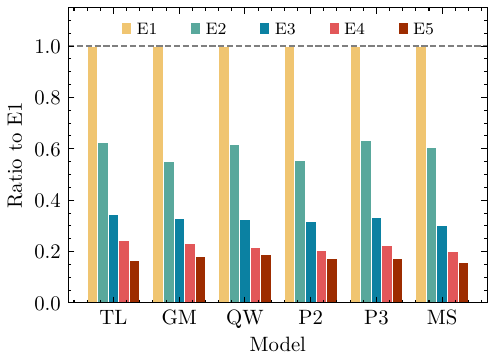}
        \subcaption{Time per token}
        \label{fig:time_per_token_bar}
    \end{minipage}
    \begin{minipage}[t]{0.32\textwidth}
        \includegraphics[width=\linewidth]{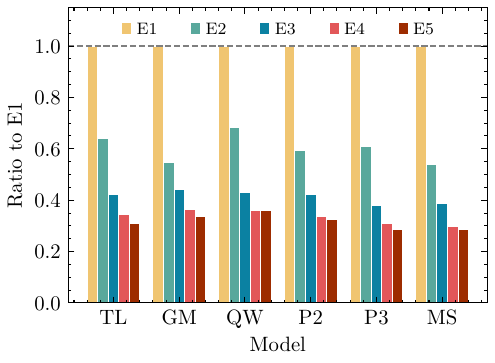}
        \subcaption{Energy per token}
        \label{fig:energy_per_token_bar}
    \end{minipage}
    \begin{minipage}[t]{0.32\textwidth}
        \includegraphics[width=\linewidth]{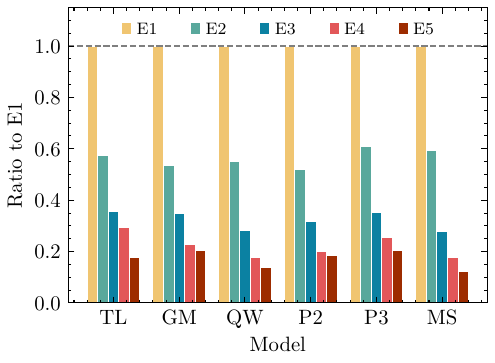}
        \subcaption{TTFT per input token}
        \label{fig:precision_bar}
    \end{minipage}
    \caption{Ratios of token-normalized time, energy, and TTFT relative to E1 across resource envelopes, averaged across benchmarks}
    \label{fig:ratio_averaged}
\end{figure*}
\begin{figure*}[tb]
  \centering
  \begin{minipage}[t]{0.32\textwidth}
    \includegraphics[width=\linewidth]{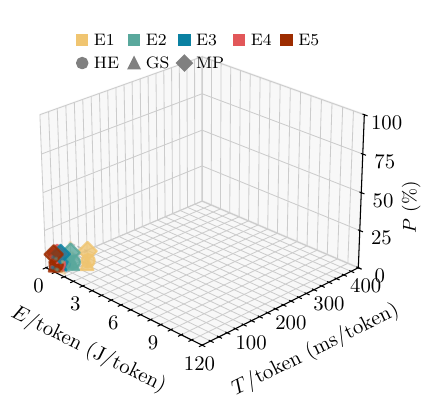}
    \subcaption{TinyLlama}
    \label{fig:tinyllama_per_token_3d_HE}
  \end{minipage}\hfill
  \begin{minipage}[t]{0.32\textwidth}
    \includegraphics[width=\linewidth]{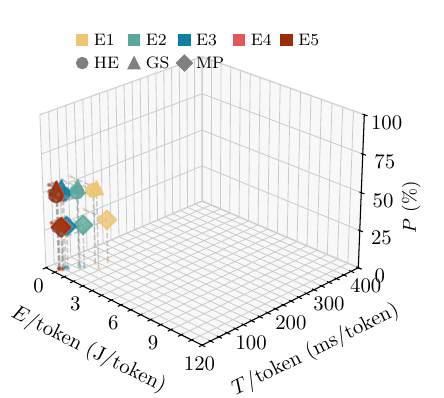}
    \subcaption{Qwen2.5}
    \label{fig:qwen2_5_per_token_3d_HE}
  \end{minipage}\hfill
  \begin{minipage}[t]{0.32\textwidth}
    \includegraphics[width=\linewidth]{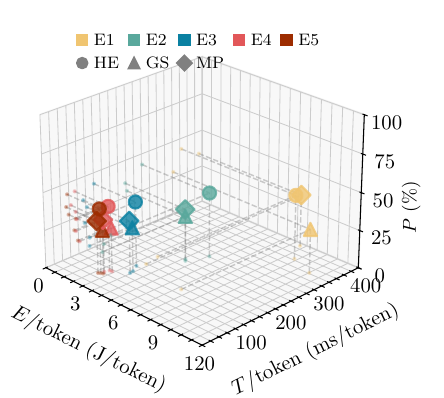}
    \subcaption{Mistral}
    \label{fig:mistral_per_token_3d_HE}
  \end{minipage}
  \caption{Precision, time per token, and energy per token 3D plots for each model under different envelopes across all benchmarks}
  \label{fig:trajectory_3d}
\end{figure*}
RQ3 examines how tightening the resource envelope across different deployment scenarios affects the measured PTME metrics, and how this effect depends on model scale and task. 

Table~\ref{tab:ptme_humaneval} reports the PTME metrics for HumanEval across envelopes E1--E5 as a representative case (other benchmarks are in the Supplementary material). Precision and the input/output token counts are completely identical across all envelopes, confirming that the deterministic decoding with temperature set to zero works as intended. The remaining metrics differ significantly across envelopes at the 1\% level according to a statistical test. The direction of these changes, however, differs by metric: time- and energy-related metrics are smaller under more resource-rich envelopes, whereas memory shows the opposite tendency, being slightly smaller under the more constrained envelopes. This is because tighter envelopes limit the available RAM and number of threads, reducing the working memory allocated by the runtime; the difference is nonetheless small compared with the footprint differences between models. Focusing on the absolute value of energy per token, it is larger under tighter envelopes. This has an important implication for edge and mobile deployment. Such environments are expected to operate under limited power and energy budgets, yet tighter resource constraints increase the energy required for the same inference task. This creates a twofold burden: the available energy budget is smaller, and the energy cost per inference increases.

Fig.~\ref{fig:ratio_averaged} shows the ratio of time per token, energy per token, and TTFT relative to the most constrained envelope (E1) averaged across all benchmarks (benchmark-specific results are in the Supplementary material). The horizontal axis shows the models, while the vertical axis shows the percentage of each metric relative to the value of E1. For most models, time per token under E5 is reduced to roughly 15--19\% of its E1 value (i.e., E1 requires about 5.3--6.3 times longer than E5). Similarly, TTFT is also reduced to roughly 12--20\% (i.e., E1 requires about 4.8--8.1 times longer than E5). In contrast, the reduction in energy per token is smaller, to about 28--36\% (i.e., E1 requires about 2.7--3.5 times more). The fact that time decreases more steeply than energy indicates that more resource-rich envelopes operate at higher average power. Thus, relaxing the envelope substantially shortens inference time while reducing energy to a lesser extent, showing that optimizing for inference time and optimizing for energy consumption do not necessarily coincide.

Fig.~\ref{fig:trajectory_3d} shows the three-dimensional trajectories of precision, time per token, and energy per token across envelopes E1--E5 for the smallest model (TinyLlama), an intermediate one (Qwen2.5), and the largest one (Mistral). Because precision is invariant across envelopes, the points of each model move only within the time--energy plane while retaining their height along the precision axis. The points of TinyLlama remain tightly clustered in the low-cost region near the origin across all envelopes, but at low precision. Qwen2.5 likewise stays in the low-cost region while retaining higher precision. Mistral, by contrast, moves toward a high-cost region far from the origin as the envelope tightens, with time and energy increasing sharply. Comparing these trajectories shows that sensitivity to the envelope depends on model scale. Larger models incur more pronounced cost increases under tighter envelopes, potentially offsetting their precision advantage with a disproportionately large cost increase.

In summary, inference efficiency is not an intrinsic property of a model alone, but an interaction between the model and the resource envelope in which it is executed. Tightening the envelope does not change task capability, but it substantially changes the physical cost required to obtain the same output. This effect is not uniform: latency is amplified more strongly than energy, and larger models are more sensitive to resource constraints than smaller ones. Therefore, a model that appears efficient or acceptable on a desktop-class computer may not remain efficient under edge-class constraints. These results indicate that the practical deployment of lightweight LLMs requires evaluation and model selection based on profiling that takes their target execution envelope into account.

\subsection{Threats to Validity and Scope of Conclusions}
\label{sec:threats_validity}

All experiments were conducted on a desktop-class machine under explicitly constrained resource envelopes rather than on actual edge or mobile hardware. The envelopes approximate edge-class resource limits by capping CPU frequency, core count, and memory, but they do not reproduce the full hardware stack of any specific device. While this setup enables controlled and reproducible PTME measurements, absolute values may differ on real devices. Our conclusions are therefore limited to relative trends, trade-offs, and sensitivity to resource constraints, and do not claim device-specific performance or energy behavior. In particular, the envelope-sensitivity results of RQ3 should be read as the response of inference cost to tightening resource limits on this platform, not as predictions for particular edge devices.

Energy was measured through the Intel RAPL interface for the CPU package domain only. It therefore excludes other system-level contributions such as DRAM, storage, or peripherals, and reflects the dominant CPU contribution under CPU-only execution rather than total device energy. In addition, since all experiments were run on CPU only, GPU-accelerated inference is not considered, despite the growing prevalence of GPU-equipped edge devices. On a GPU, time and energy would differ substantially owing to the higher parallelism and power profile of the accelerator. Memory behavior would also change: model weights and activations would reside in GPU memory rather than host RAM, and the available GPU memory would bound the model sizes that can be loaded.

Inference was executed using a single backend with fixed decoding parameters. This ensures determinism and comparability across models and envelopes, but limits generalization to other inference runtimes or decoding strategies. In addition, the set of evaluated models and workloads is necessarily finite, and the results may not generalize to substantially larger models or fundamentally different tasks. Despite these limitations, the experimental design is sufficient to answer RQ1--RQ3 within the stated scope.

\section{Conclusion and Future Work}
\label{sec:conclusion}
This paper introduced a PTME framework for the precision-aware profiling of lightweight LLM inference, jointly measuring precision, time, memory, and energy through direct hardware-level measurements. The framework was applied to a representative set of lightweight LLMs executed locally on a constrained desktop platform that approximates edge-class resource envelopes, using benchmarks spanning code generation, mathematical reasoning, and multi-task understanding.

Across three research questions, our experiments yield the following findings. First, commonly used static proxies approximate inference cost well but fail to predict precision. Second, under a fixed envelope, no single model dominates across all PTME dimensions: a four-objective Pareto analysis reveals non-dominated configurations that single-metric evaluation overlooks, and the most balanced choice is a mid-sized model rather than the largest or the smallest. Third, tightening the envelope increases cost without affecting precision, amplifying inference time more strongly than energy and penalizing larger models the most.

In practical terms, Pareto analysis should be used not to identify a universal best model, but to remove dominated choices and identify non-dominated configurations whose precision gains justify their physical cost. For edge-class deployment, such non-dominated configurations are often more useful than either the smallest or the largest model, because they balance usability and resource feasibility. For  desktop-class deployments, the Pareto view clarifies when the additional precision of a larger model is worth the extra time, memory, and energy.

Taken together, these results show that proxy- or single-metric assessments are insufficient for comparing lightweight LLMs, and that the practical cost of local inference depends not only on the model but also on how strongly the execution environment constrains computation. Meaningful comparison, therefore, requires direct PTME evaluation that jointly accounts for correctness and physical cost, providing a basis for sustainable LLM deployment under resource constraints.

This framework opens several directions for future work, including energy-aware and envelope-aware model selection, adaptive inference that responds to the available resource budget, the design of models and inference configurations suited to resource-constrained environments, and the extension to multi-agent LLM systems under strict resource constraints.

\section*{Acknowledgements}
The authors used ChatGPT (GPT-5.5) and Claude (Opus 4.8) to assist in drafting and refining the English text of this paper. All AI-generated text was reviewed, edited, and verified by the authors, who take full responsibility for the content of this paper.

\balance
\bibliographystyle{IEEEtran}

\bibliography{references}

@inproceedings{alba2026software,
  author    = {Enrique Alba and Hector D. Menendez},
  title     = {White-Box Execution Refactoring of Transformers for Lower Energy},
  booktitle = {Search-Based Software Engineering},
  series    = {Lecture Notes in Computer Science},
  publisher = {Springer},
  year      = {2026},
  note      = {{I}n press, SSBSE 2026 SSBSE Challenge track},
  uurl       = {https://conf.researchr.org/details/ssbse-2026/ssbse-2026-ssbse-challenge/1/White-Box-Execution-Refactoring-of-Transformers-for-Lower-Energy}
}

@inproceedings{strubell2019energy,
    title = "{Energy and Policy Considerations for Deep Learning in {NLP}}",
    author = "Strubell, Emma  and
      Ganesh, Ananya  and
      McCallum, Andrew",
    editor = "Korhonen, Anna  and
      Traum, David  and
      M{\`a}rquez, Llu{\'i}s",
    booktitle = "Proceedings of the 57th Annual Meeting of the Association for Computational Linguistics",
    month = jul,
    year = "2019",
    address = "Florence, Italy",
    publisher = "Association for Computational Linguistics",
    uurl = "https://aclanthology.org/P19-1355/",
    doi = "10.18653/v1/P19-1355",
    pages = "3645--3650"
}

@article{schwartz2020green,
author = {Schwartz, Roy and Dodge, Jesse and Smith, Noah A. and Etzioni, Oren},
title = {{Green AI}},
year = {2020},
issue_date = {December 2020},
publisher = {Association for Computing Machinery},
address = {New York, NY, USA},
volume = {63},
number = {12},
issn = {0001-0782},
uurl = {https://doi.org/10.1145/3381831},
doi = {10.1145/3381831},
journal = {Commun. ACM},
month = nov,
pages = {54–63},
numpages = {10}
}

@article{shi2016edge,
  author={Shi, Weisong and Cao, Jie and Zhang, Quan and Li, Youhuizi and Xu, Lanyu},
  journal={IEEE Internet of Things Journal}, 
  title={{Edge Computing: Vision and Challenges}}, 
  year={2016},
  volume={3},
  number={5},
  pages={637-646},
  uurl={https://doi.org/10.1109/JIOT.2016.2579198},
  doi={10.1109/JIOT.2016.2579198}}

@article{satyanarayanan2017edge,
  author={Satyanarayanan, Mahadev},
  journal={Computer}, 
  title={{The Emergence of Edge Computing}}, 
  year={2017},
  volume={50},
  number={1},
  pages={30-39},
  uurl={https://doi.org/10.1109/MC.2017.9},
  doi={10.1109/MC.2017.9}}

@inproceedings{horowitz2014energy,
  author={Horowitz, Mark},
  booktitle={2014 IEEE International Solid-State Circuits Conference Digest of Technical Papers (ISSCC)}, 
  title={{1.1 Computing's energy problem (and what we can do about it)}}, 
  year={2014},
  volume={},
  number={},
  pages={10-14},
  uurl={https://doi.org/10.1109/ISSCC.2014.6757323},
  doi={10.1109/ISSCC.2014.6757323}}

@inproceedings{dettmers2022llm,
 author = {Dettmers, Tim and Lewis, Mike and Belkada, Younes and Zettlemoyer, Luke},
 booktitle = {Advances in Neural Information Processing Systems},
 editor = {S. Koyejo and S. Mohamed and A. Agarwal and D. Belgrave and K. Cho and A. Oh},
 pages = {30318--30332},
 publisher = {Curran Associates, Inc.},
 title = {{LLM.int8(): 8-bit Matrix Multiplication for Transformers at Scale}},
 uurl = {https://proceedings.neurips.cc/paper_files/paper/2022/file/c3ba4962c05c49636d4c6206a97e9c8a-Paper-Conference.pdf},
 volume = {35},
 year = {2022}
}

@misc{frantar2023gptq,
      title={{GPTQ: Accurate Post-Training Quantization for Generative Pre-trained Transformers}}, 
      author={Elias Frantar and Saleh Ashkboos and Torsten Hoefler and Dan Alistarh},
      year={2023},
      eprint={2210.17323},
      archivePrefix={arXiv},
      primaryClass={cs.LG},
      url={https://arxiv.org/abs/2210.17323}, 
}

@article{varghese2021survey,
author = {Varghese, Blesson and Wang, Nan and Bermbach, David and Hong, Cheol-Ho and Lara, Eyal De and Shi, Weisong and Stewart, Christopher},
title = {{A Survey on Edge Performance Benchmarking}},
year = {2021},
issue_date = {April 2022},
publisher = {Association for Computing Machinery},
address = {New York, NY, USA},
volume = {54},
number = {3},
issn = {0360-0300},
uurl = {https://doi.org/10.1145/3444692},
doi = {10.1145/3444692},
journal = {ACM Computing Surveys},
month = apr,
articleno = {66},
numpages = {33},
pages = {1--33}
}

@article{ordonez2025intelligent,
AUTHOR = {Cajas Ordo\~nez, Sebasti\'an A. and Samanta, Jaydeep and Su\'arez-Cetrulo, Andr\'es L. and Carbajo, Ricardo Sim\'on},
TITLE = {{Intelligent Edge Computing and Machine Learning: A Survey of Optimization and Applications}},
JOURNAL = {Future Internet},
VOLUME = {17},
YEAR = {2025},
NUMBER = {9},
ARTICLE-NUMBER = {417},
uURL = {https://www.mdpi.com/1999-5903/17/9/417},
ISSN = {1999-5903},
DOI = {10.3390/fi17090417}
}

@misc{touvron2023llama,
      title={{LLaMA: Open and Efficient Foundation Language Models}}, 
      author={Hugo Touvron and Thibaut Lavril and Gautier Izacard and Xavier Martinet and Marie-Anne Lachaux and Timothée Lacroix and Baptiste Rozière and Naman Goyal and Eric Hambro and Faisal Azhar and Aurelien Rodriguez and Armand Joulin and Edouard Grave and Guillaume Lample},
      year={2023},
      eprint={2302.13971},
      archivePrefix={arXiv},
      primaryClass={cs.CL},
      url={https://arxiv.org/abs/2302.13971}, 
}

@misc{jiang2023mistral,
      title={{Mistral 7B}}, 
      author={Albert Q. Jiang and Alexandre Sablayrolles and others},
      year={2023},
      eprint={2310.06825},
      archivePrefix={arXiv},
      primaryClass={cs.CL},
      url={https://arxiv.org/abs/2310.06825}, 
}

@misc{zhang2024tinyllama,
      title={{TinyLlama: An Open-Source Small Language Model}}, 
      author={Peiyuan Zhang and Guangtao Zeng and Tianduo Wang and Wei Lu},
      year={2024},
      eprint={2401.02385},
      archivePrefix={arXiv},
      primaryClass={cs.CL},
      url={https://arxiv.org/abs/2401.02385}, 
}

@misc{phi2_microsoft,
  author       = {{Microsoft Research}},
  title        = {{Phi-2: The Surprising Power of Small Language Models}},
  year         = {2023},
  howpublished = {\url{https://www.microsoft.com/en-us/research/blog/phi-2-the-surprising-power-of-small-language-models/}},
  note         = {Accessed: 2026-06-08}
}

@misc{abdin2024phi3,
      title={{Phi-3 Technical Report: A Highly Capable Language Model Locally on Your Phone}}, 
      author={Marah Abdin and Jyoti Aneja and Hany Awadalla and Ahmed Awadallah and Ammar Ahmad Awan and others},
      year={2024},
      eprint={2404.14219},
      archivePrefix={arXiv},
      primaryClass={cs.CL},
      url={https://arxiv.org/abs/2404.14219}, 
}

@misc{qwen2025qwen25technicalreport,
      title={{Qwen2.5 Technical Report}}, 
      author={{Qwen Team}},
      year={2025},
      eprint={2412.15115},
      archivePrefix={arXiv},
      primaryClass={cs.CL},
      url={https://arxiv.org/abs/2412.15115}, 
}

@misc{gemma2024technical,
      title={Gemma: Open Models Based on Gemini Research and Technology}, 
      author={{Gemma Team}},
      year={2024},
      eprint={2403.08295},
      archivePrefix={arXiv},
      primaryClass={cs.CL},
      url={https://arxiv.org/abs/2403.08295}, 
}

@article{henderson2020sustainable,
  author  = {Peter Henderson and Jieru Hu and Joshua Romoff and Emma Brunskill and Dan Jurafsky and Joelle Pineau},
  title   = {{Towards the Systematic Reporting of the Energy and Carbon Footprints of Machine Learning}},
  journal = {Journal of Machine Learning Research},
  year    = {2020},
  volume  = {21},
  number  = {248},
  pages   = {1--43},
  uurl     = {http://jmlr.org/papers/v21/20-312.html}
}

@inproceedings{Polo2024tinyBenchmarks,
title={tinyBenchmarks: evaluating {LLM}s with fewer examples},
author={Felipe Maia Polo and Lucas Weber and Leshem Choshen and Yuekai Sun and Gongjun Xu and Mikhail Yurochkin},
booktitle={Forty-first International Conference on Machine Learning},
year={2024},
uurl={https://openreview.net/forum?id=qAml3FpfhG}
}

@inproceedings{Wang2024MMLU-Pro,
 author = {Wang, Yubo and Ma, Xueguang and Zhang, Ge and Ni and others},
 booktitle = {Advances in Neural Information Processing Systems},
 doi = {10.52202/079017-3018},
 editor = {A. Globerson and L. Mackey and D. Belgrave and A. Fan and U. Paquet and J. Tomczak and C. Zhang},
 pages = {95266--95290},
 publisher = {Curran Associates, Inc.},
 title = {{MMLU-Pro: A More Robust and Challenging Multi-Task Language Understanding Benchmark}},
 uurl = {https://proceedings.neurips.cc/paper_files/paper/2024/file/ad236edc564f3e3156e1b2feafb99a24-Paper-Datasets_and_Benchmarks_Track.pdf},
 volume = {37},
 year = {2024}
}

@inproceedings{
hendrycks2021MMLU,
title={{Measuring Massive Multitask Language Understanding}},
author={Dan Hendrycks and Collin Burns and Steven Basart and Andy Zou and Mantas Mazeika and Dawn Song and Jacob Steinhardt},
booktitle={International Conference on Learning Representations},
year={2021},
uurl={https://openreview.net/forum?id=d7KBjmI3GmQ}
}

@misc{cobbe2021trainingverifierssolvemath,
      title={{Training Verifiers to Solve Math Word Problems}}, 
      author={Karl Cobbe and Vineet Kosaraju and Mohammad Bavarian and Mark Chen and Heewoo Jun and Lukasz Kaiser and Matthias Plappert and Jerry Tworek and Jacob Hilton and Reiichiro Nakano and Christopher Hesse and John Schulman},
      year={2021},
      eprint={2110.14168},
      archivePrefix={arXiv},
      primaryClass={cs.LG},
      url={https://arxiv.org/abs/2110.14168}, 
}

@misc{chen2021evaluatinglargelanguagemodels,
      title={{Evaluating Large Language Models Trained on Code}}, 
      author={Mark Chen and Jerry Tworek and Heewoo Jun and others},
      year={2021},
      eprint={2107.03374},
      archivePrefix={arXiv},
      primaryClass={cs.LG},
      url={https://arxiv.org/abs/2107.03374}, 
}

@ARTICLE{Abadade2023TinyML,

  author={Abadade, Youssef and Temouden, Anas and Bamoumen, Hatim and Benamar, Nabil and Chtouki, Yousra and Hafid, Abdelhakim Senhaji},

  journal={IEEE Access},

  title={{A Comprehensive Survey on TinyML}},

  year={2023},

  volume={11},

  number={},

  pages={96892-96922},

  uurl={https://doi.org/10.1109/ACCESS.2023.3294111},
  doi={10.1109/ACCESS.2023.3294111}}

@inproceedings{NEURIPS2023_44956951,
 author = {Ma, Xinyin and Fang, Gongfan and Wang, Xinchao},
 booktitle = {Advances in Neural Information Processing Systems},
 editor = {A. Oh and T. Naumann and A. Globerson and K. Saenko and M. Hardt and S. Levine},
 pages = {21702--21720},
 publisher = {Curran Associates, Inc.},
 title = {{LLM-Pruner: On the Structural Pruning of Large Language Models}},
 uurl = {https://proceedings.neurips.cc/paper_files/paper/2023/file/44956951349095f74492a5471128a7e0-Paper-Conference.pdf},
 volume = {36},
 year = {2023}
}

@misc{thop,
  author       = {{Ultralytics}},
  title        = {{THOP: PyTorch-OpCounter}},
  howpublished = {\url{https://github.com/ultralytics/thop}},
  year         = {2025},
  note         = {Accessed: 2026-06-08}
}

@misc{xu2024ondevice,
      title={{On-Device Language Models: A Comprehensive Review}}, 
      author={Jiajun Xu and Zhiyuan Li and Wei Chen and Qun Wang and Xin Gao and Qi Cai and Ziyuan Ling},
      year={2024},
      eprint={2409.00088},
      archivePrefix={arXiv},
      primaryClass={cs.CL},
      url={https://arxiv.org/abs/2409.00088}, 
}

@article{KIBRIYA2024109698,
title = {{Privacy issues in Large Language Models: A survey}},
journal = {Computers and Electrical Engineering},
volume = {120},
pages = {109698},
year = {2024},
issn = {0045-7906},
doi = {https://doi.org/10.1016/j.compeleceng.2024.109698},
uurl = {https://www.sciencedirect.com/science/article/pii/S0045790624006256},
author = {Hareem Kibriya and Wazir Zada Khan and Ayesha Siddiqa and Muhammad Khurram Khan}
}

@article{Zheng2025Review,
author = {Zheng, Yue and Chen, Yuhao and Qian, Bin and Shi, Xiufang and Shu, Yuanchao and Chen, Jiming},
title = {{A Review on Edge Large Language Models: Design, Execution, and Applications}},
year = {2025},
issue_date = {August 2025},
publisher = {Association for Computing Machinery},
address = {New York, NY, USA},
volume = {57},
number = {8},
issn = {0360-0300},
uurl = {https://doi.org/10.1145/3719664},
doi = {10.1145/3719664},
journal = {ACM Computing Surveys},
month = mar,
articleno = {209},
numpages = {35}
}

@article{HARADA2026108482,
title = {{Green optimization: Energy-aware design of metaheuristics by using machine learning surrogates to cope with real problems}},
journal = {Future Generation Computer Systems},
volume = {182},
pages = {108482},
year = {2026},
issn = {0167-739X},
doi = {https://doi.org/10.1016/j.future.2026.108482},
uurl = {https://www.sciencedirect.com/science/article/pii/S0167739X26001160},
author = {Tomohiro Harada and Enrique Alba and Gabriel Luque}
}

@inproceedings{alba2026energyawaremetaheuristics,
author = {Alba, Enrique and Harada, Tomohiro and Luque, Gabriel},
title = {Energy-Aware Metaheuristics},
year = {2026},
isbn = {9798400724879},
publisher = {Association for Computing Machinery},
address = {New York, NY, USA},
uurl = {https://doi.org/10.1145/3795095.3805196},
doi = {10.1145/3795095.3805196},
booktitle = {Proceedings of the Genetic and Evolutionary Computation Conference},
pages = {672–680},
numpages = {9},
location = {Centro Internacional de Convenciones CIC-ANDE, San Jose, Costa Rica},
series = {GECCO '26}
}

\includepdf[pages=-]{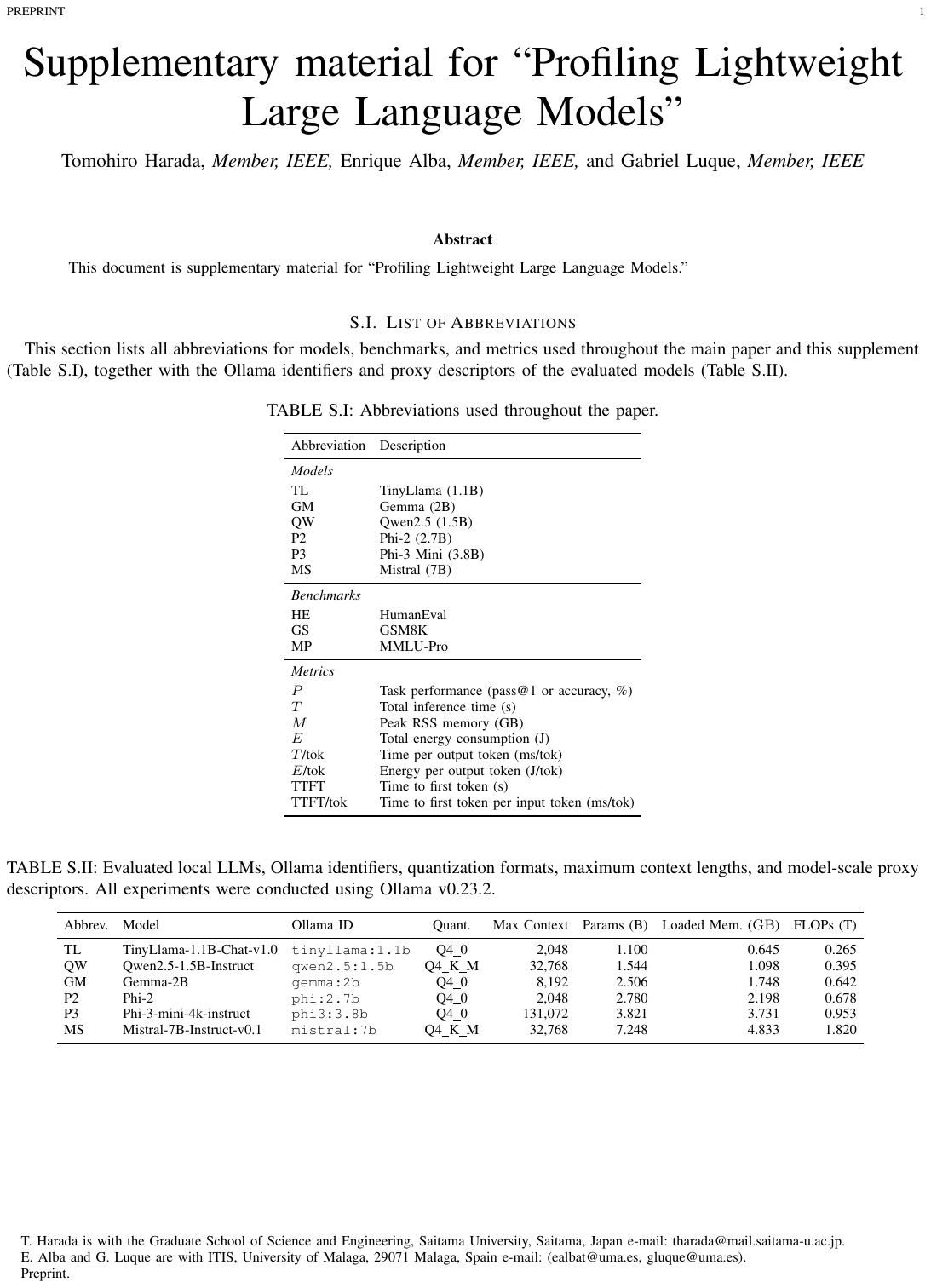}
\end{document}